%% file: FastNeRF-main.tex
\documentclass[10pt,twocolumn,letterpaper]{article}
\usepackage[accsupp]{axessibility}
\usepackage[pagenumbers]{cvpr} 

\usepackage{graphicx}
\usepackage{amsmath}
\usepackage{amssymb}
\usepackage{booktabs}
\usepackage{cuted}
\usepackage[table,xcdraw,dvipsnames]{xcolor}
\usepackage{textcomp}
\usepackage{tabularx}
\usepackage[numbers,sort]{natbib}
\usepackage{multirow}
\usepackage{microtype}
\usepackage{wrapfig}
\usepackage{MnSymbol,bbding,pifont}
\usepackage{graphbox}
\graphicspath{{figures/}}

\makeatletter
\renewcommand\paragraph{\@startsection{paragraph}{4}{\z@}%
	{0.75ex \@plus.5ex \@minus.2ex}%
	{-1em}%
	{\normalfont\normalsize\bfseries\maybe@addperiod}}
\newcommand{\maybe@addperiod}[1]{#1\@addpunct{.}}
\makeatother

\usepackage[breaklinks,colorlinks,bookmarks=false,linkcolor=blue,citecolor=blue]{hyperref}
\usepackage[nameinlink,noabbrev,capitalise]{cleveref}
\crefformat{equation}{#2Equation~#1#3}
\newcommand\rurl[1]{%
  \href{http://#1}{\nolinkurl{#1}}%
}

\begin{document}

\title{Learning Neural Duplex Radiance Fields for Real-Time View Synthesis}



\author{Ziyu Wan$^{1}$ \quad 
Christian Richardt$^{2}$ \quad  Aljaž Božič$^2$  \quad Chao Li$^{2}$ \quad Vijay Rengarajan$^{2}$ \\ \quad Seonghyeon Nam$^{2}$ \quad Xiaoyu Xiang$^{2}$ \quad Tuotuo Li$^{2}$ \quad Bo Zhu$^{2}$ \quad Rakesh Ranjan$^{2}$ \quad Jing Liao$^{1}$\footnotemark   \\
	$^1$City University of Hong Kong \quad \quad \quad 
	$^2$Meta Reality Labs
	\\
        \textbf{\rurl{raywzy.com/NDRF/}}
	}



\twocolumn[{
\renewcommand\twocolumn[1][]{#1}
\maketitle
    \vspace{-2.5em}
    \setlength\tabcolsep{0.5pt}
    \centering
    \includegraphics[width=0.99\textwidth]{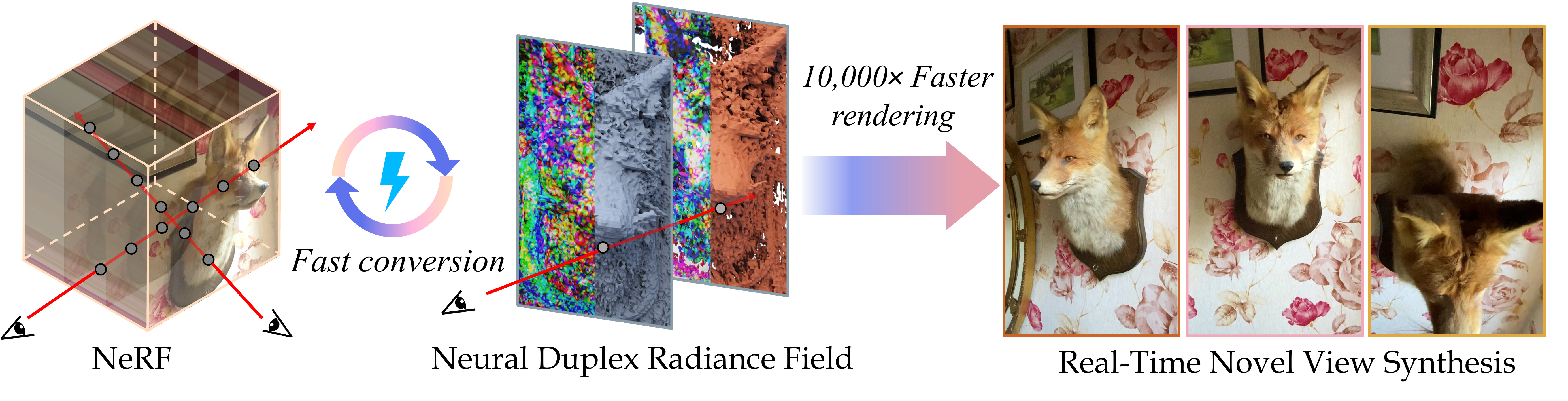}
    \vspace{-1.0em}
    \captionsetup{type=figure} 
    \captionof{figure}{\label{fig:teaser}%
        Our framework efficiently learns the neural duplex radiance field from a NeRF model for high-quality real-time view synthesis.
        %
        %
    }
    \vspace{0.5em}
}]

\maketitle
\begin{NoHyper}\footnotetext{\hspace{-1.25em}* Corresponding author.}\end{NoHyper}


\begin{abstract}

%
\noindent
Neural radiance fields (NeRFs) enable novel-view synthesis with unprecedented visual quality.
However, to render photorealistic images, NeRFs require hundreds of deep multilayer perceptron (MLP) evaluations – for each pixel.
This is prohibitively expensive and makes real-time rendering infeasible, even on powerful modern GPUs.
In this paper, we propose a novel approach to distill and bake NeRFs into highly efficient mesh-based neural representations that are fully compatible with the massively parallel graphics rendering pipeline.
We represent scenes as neural radiance features encoded on a two-layer \emph{duplex} mesh, which effectively overcomes the inherent inaccuracies in 3D surface reconstruction by learning the aggregated radiance information from a reliable interval of ray-surface intersections.
To exploit local geometric relationships of nearby pixels, we leverage screen-space convolutions instead of the MLPs used in NeRFs to achieve high-quality appearance.
Finally, the performance of the whole framework is further boosted by a novel multi-view distillation optimization strategy.
We demonstrate the effectiveness and superiority of our approach via extensive experiments on a range of standard datasets.

\end{abstract}

\input{1-introduction}
\input{2-relatedwork}

\input{3-method}
\input{4-experiments}
\input{5-conclusion}

{\small
\bibliographystyle{ieeenat_fullname}
\bibliography{egbib,FastNeRF-CR}
}

\newpage

\onecolumn

\appendix

\vspace{-2cm}
\begin{center}
\Large\textbf{Learning Neural Duplex Radiance Fields for Real-Time View Synthesis} \\ \Large\textbf{Supplementary Material}
\end{center}
\enlargethispage{\baselineskip}

\section{Overview}

In this supplemental material,  additional implementation details and experimental results are provided, including:
\begin{itemize}
  \setlength\itemsep{0em}
  \item More details about network architecture (Section \ref{Network});
  \item More details about shader implementation (Section \ref{Shader});
  
  
  \item More ablation studies about threshold selection, the teacher model and the number of mesh layers. We also provide another perspective to visualize and understand the proposed method (Section \ref{Ablation});
  \item Video demo. Please refer to our project page.
\end{itemize}

\section{Network Architecture}\label{Network}

We present the details of the two versions of our shading CNN in \cref{table:CNN}.
To generate high-frequency textures and view-dependent effects, we also leverage the positional encoding, which maps view direction and intersection locations into a higher-dimensional space.
For the CUDA version, both the position and view encoding dimensions are set to 10.
For the WebGL version, we set the dimension of view direction encoding to 5 and the others to 0.
To ensure the CNN efficiency of the CUDA version, we use depthwise separable convolution \cite{chollet2017xception} to avoid expensive computations. We also tried to implement depthwise separable convolution of WebGL version into the fragment shader but didn't attain effective speedup.

\begin{table}[!h]
\vspace{1em}
\caption{\label{table:CNN}\textbf{Detailed architecture of convolutional shading network.}}
\small
\centering
\setlength{\tabcolsep}{4.0mm}
\begin{tabular}{lcccc}
\toprule
\textbf{Version}                      & \textbf{Layer}    & \textbf{Kernel size / stride} & \textbf{Channels} & \textbf{Non-Linearity} \\ \midrule
\multirow{3}{*}{CUDA}
                            & 2DConv     & $3\times3 / (1,1)$            &  256 $\rightarrow $ 256   & $\mathtt{ReLU}$        \\
                            & 2DConv     & $3\times3 / (1,1)$            &  256 $\rightarrow $ 256   & $\mathtt{ReLU}$                  \\
                            & 2DConv     & $1\times1 / (1,1)$            &  256 $\rightarrow $ 3    & $\mathtt{Sigmoid}$                 \\     \midrule
 \multirow{2}{*}{WebGL}
                            & 2DConv     & $2\times2 / (1,1)$            &  55 $\rightarrow $ 32   & $\mathtt{ReLU}$        \\
                            & 2DConv     & $2\times2 / (1,1)$            &  32 $\rightarrow $ 3   & $\mathtt{Sigmoid}$        \\
 
                            \bottomrule
\end{tabular}
\end{table}

\section{Shader Implementation}\label{Shader}

Since the learnable features are attached on the mesh vertices, we directly generate the screen-space feature buffer using traditional hardware rasterization.
More specifically, we render 4 RGBA buffers for each mesh: 2 for 8-channel features, 1 for ray-surface intersection positions, and 1 for view directions.
To implement the convolution operator in the same framework, after optimizing the whole parameters using PyTorch, we import the 2-layer convolution weights into 2 RGBA \texttt{GL\_TEXTURE\_2D} with shape $\texttt{out\_channel} \times \texttt{in\_channel}$ thanks to the 2$\times$2 spatial kernel.
We employ two passes in total to generate view-dependent RGB frames, one pass for each convolutional layer.
In each pass, we only consider the radiance information of a local receptive field, which can be efficiently queried through \texttt{texelFetch}.
To minimize the memory footprint, we implement \texttt{sin/cos} positional encoding in the fragment shader of the first convolution layer rather than in the rasterization step.
We additionally tried to put the forward propagation of all CNN layers in one fragment shader, but found the efficiency to be poor.
There are also some well-known existing web-based deep learning frameworks like \texttt{TF}.\texttt{js}, but we found they could not efficiently support the high-resolution synthesis.


\section{Additional Ablation Studies}\label{Ablation}

\paragraph{Results of Single Surface using Different Thresholds}

In the main paper, we have conducted ablation studies on the effectiveness of duplex radiance fields by using a single extracted mesh with a threshold $10^{-4}$.
Here another interesting question is, what will be the best performance if we adjust the threshold to get a better surface?
Will this surpass the performance of neural duplex radiance fields?
To answer these questions, we conducted both quantitative and qualitative experiments on NeRF-Synthetic Ficus  data \cite{mildenhall2020nerf}.
As shown in \cref{tab:quantitative_threshold} and \cref{fig:comp_threshold}, carefully fine-tuning the threshold will lead to some performance improvements, but the rendering quality is still not acceptable due to the inaccurate geometry.
In contrast, through learning the aggregation of radiance features on two different coarse geometry surfaces, our method achieves significantly better novel-view synthesis quality than all these variations. Please note the isolated parts of extracted mesh will be automatically removed with respect to their diameters.

\begin{table*}
\caption{\label{tab:quantitative_threshold}%
   \textbf{Quantitative comparisons on Synthetic-NeRF \cite{mildenhall2020nerf} Ficus data with different thresholds.} Our approach significantly outperforms single-surface based baselines.
}
\centering
\setlength{\tabcolsep}{4.0mm}
\begin{tabular}{lcccccc}
    \toprule
        & $10^{-4}$ &$5 \times 10^{-4}$ & $10^{-3}$ & $5 \times 10^{-3}$ &$10^{-2}$ & \bf Ours \\
    \midrule
    PSNR $\uparrow$    &   25.50   &   26.91   &   27.52   &   28.01   &   27.05   & \bf 32.67 \\
    SSIM $\uparrow$    &   0.921   &   0.936   &   0.941   &   0.944   &   0.935   & \bf 0.975 \\
    LPIPS $\downarrow$ &   0.088   &   0.073   &   0.068   &   0.063   &   0.069   & \bf 0.024 \\
    \bottomrule
\end{tabular}
\end{table*}

\begin{table*}
\caption{\label{tab: different_teacher}%
   \textbf{ Quantitative comparisons using different teacher models.}
  On Synthetic-NeRF \cite{mildenhall2020nerf} Chair data, we show our method have generalization capabilities to different NeRF models.
}
\centering
\setlength{\tabcolsep}{4.0mm}
\begin{tabular}{lcccc}
    \toprule
        & TensoRF~\cite{Chen2022ECCV} & Ours-TensoRF & Instant-NGP~\cite{mueller2022instant} & Ours-Instant-NGP \\
    \midrule
    PSNR $\uparrow$  & \textbf{34.73}   & 34.08        & 34.30       & 34.17            \\
    SSIM $\uparrow$   & 0.981   & \textbf{0.983}        & 0.979       & 0.982            \\
    LPIPS $\downarrow$ & 0.013   & 0.013        & \textbf{0.010}       & 0.011           \\
    \bottomrule
\end{tabular}
\end{table*}

\begin{figure*}[tbp]
\centering
\setlength{\tabcolsep}{1pt}
\begin{tabular}{ccccccc}

\rotatebox[origin=c]{90}{Mesh \strut} & 
\includegraphics[align=c,width=0.156\textwidth]{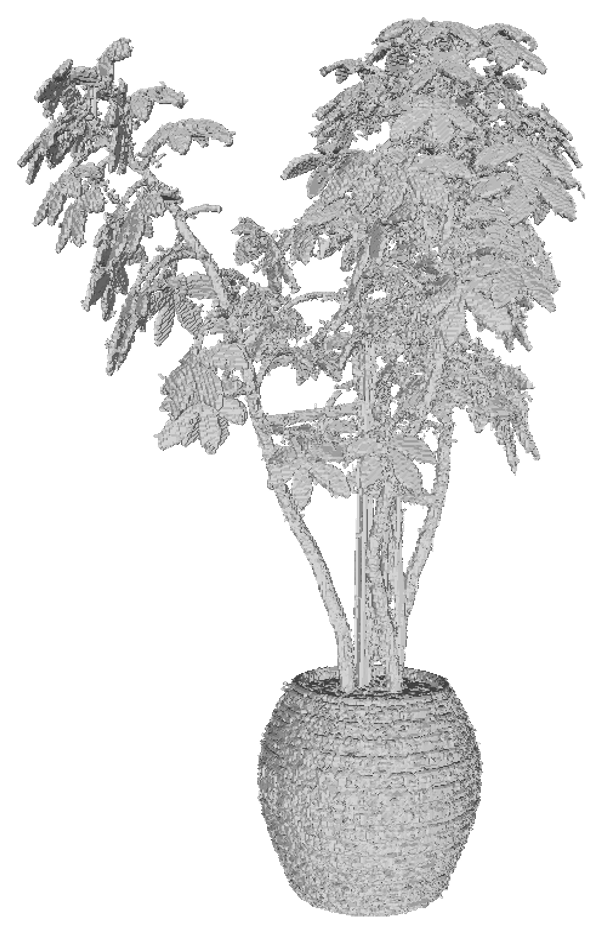} & \includegraphics[align=c,width=0.156\textwidth]{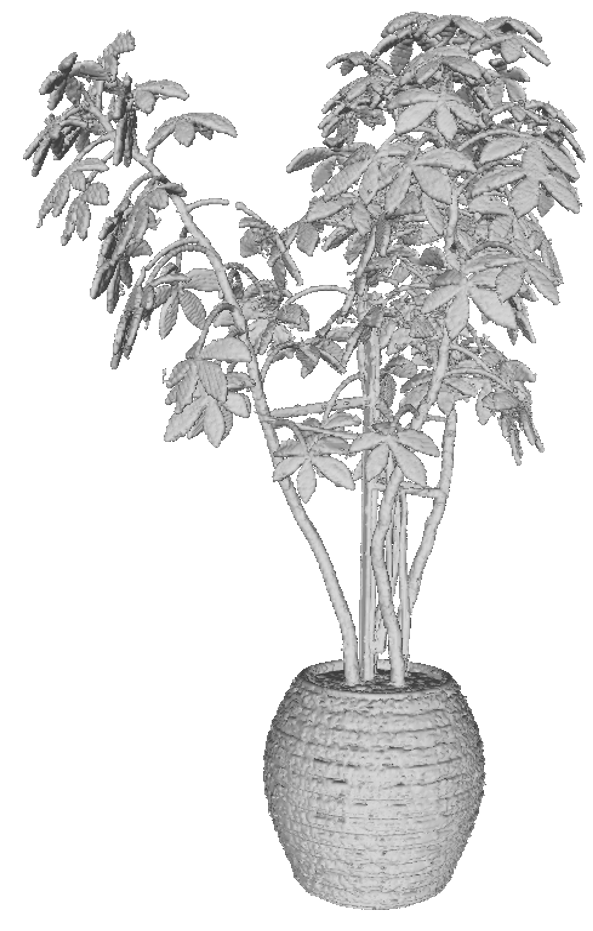} & \includegraphics[align=c,width=0.156\textwidth]{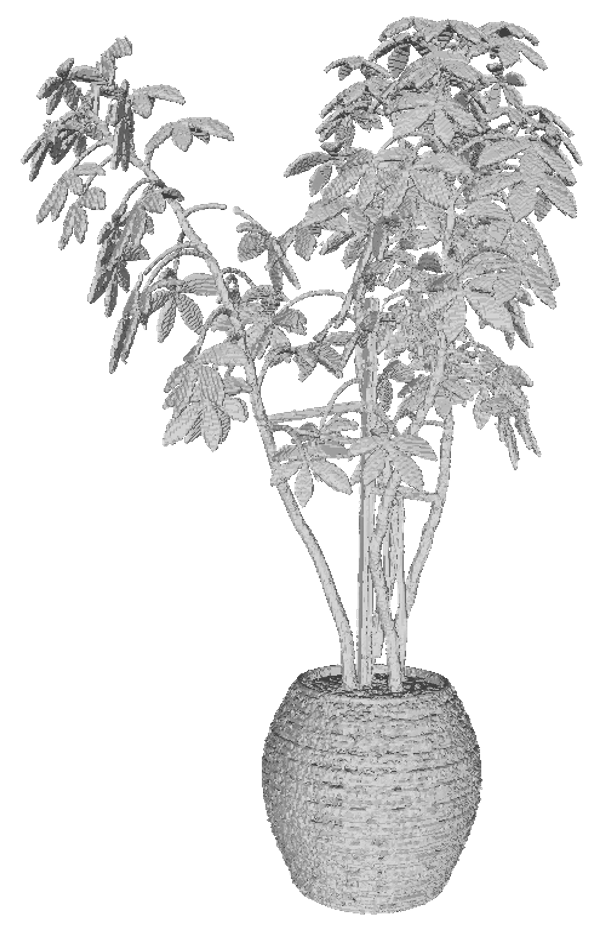} & \includegraphics[align=c,width=0.156\textwidth]{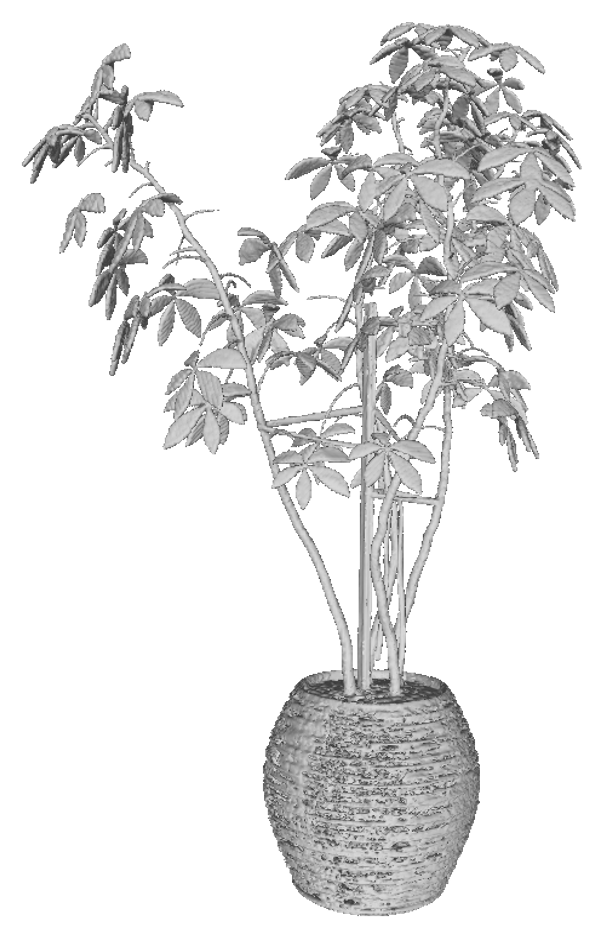} &
\includegraphics[align=c,width=0.156\textwidth]{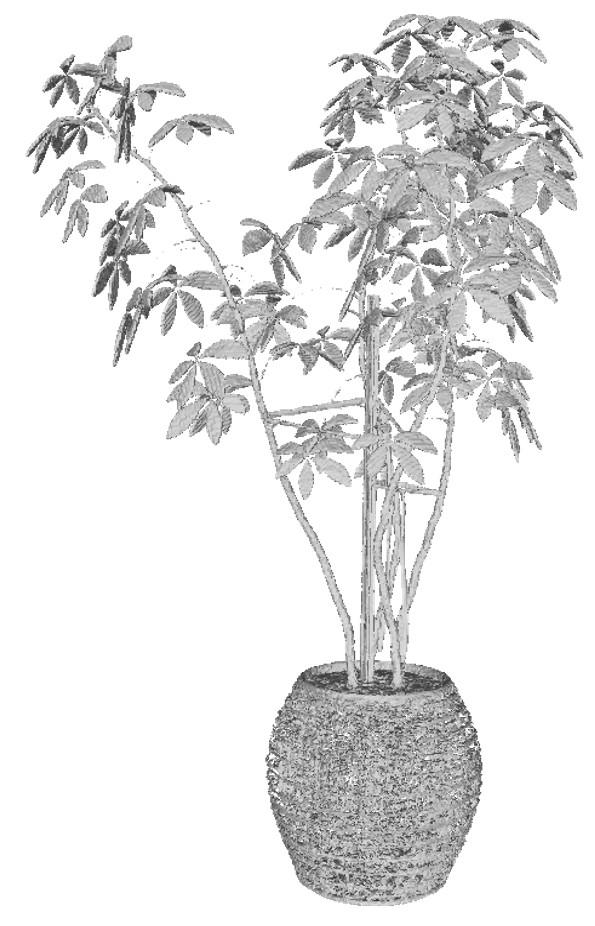}\\

\rotatebox[origin=c]{90}{Rendering \strut} & 
\includegraphics[align=c,width=0.156\textwidth]{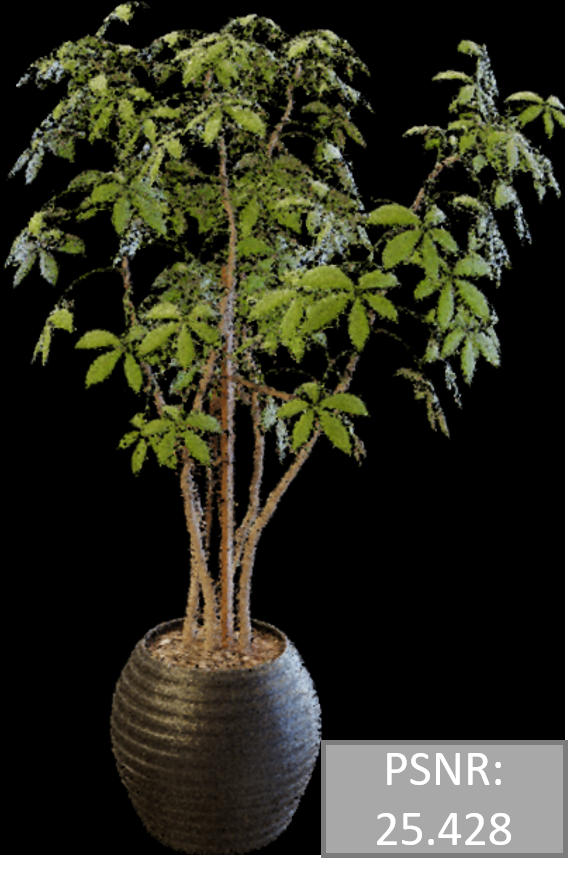} & \includegraphics[align=c,width=0.156\textwidth]{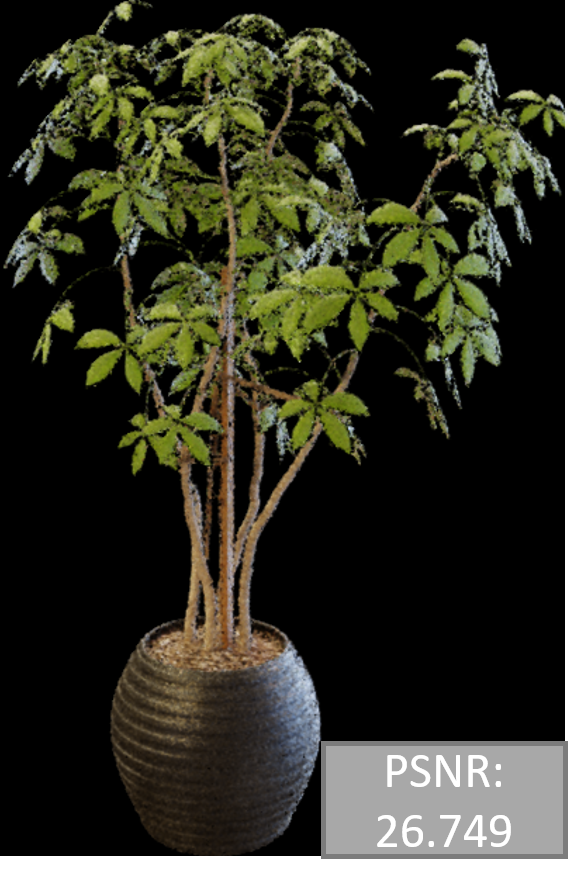} & \includegraphics[align=c,width=0.156\textwidth]{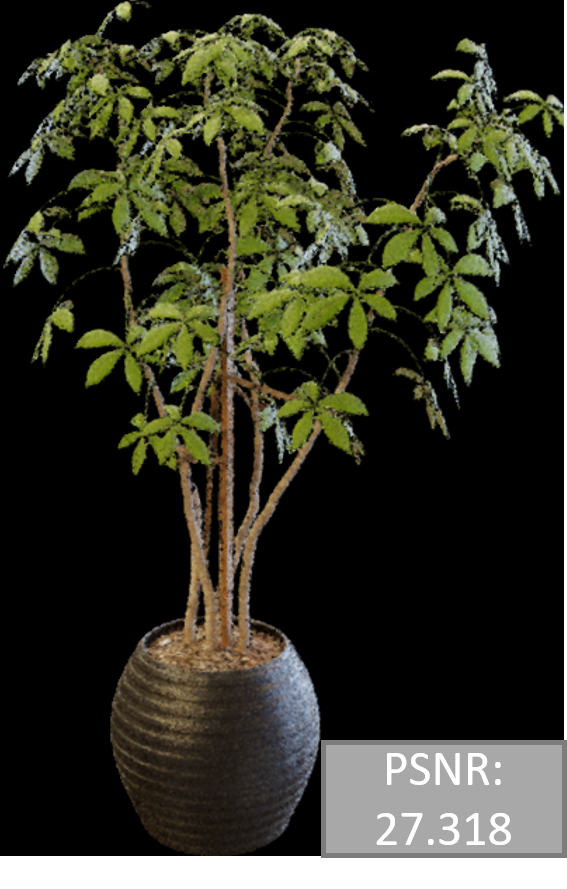} & \includegraphics[align=c,width=0.156\textwidth]{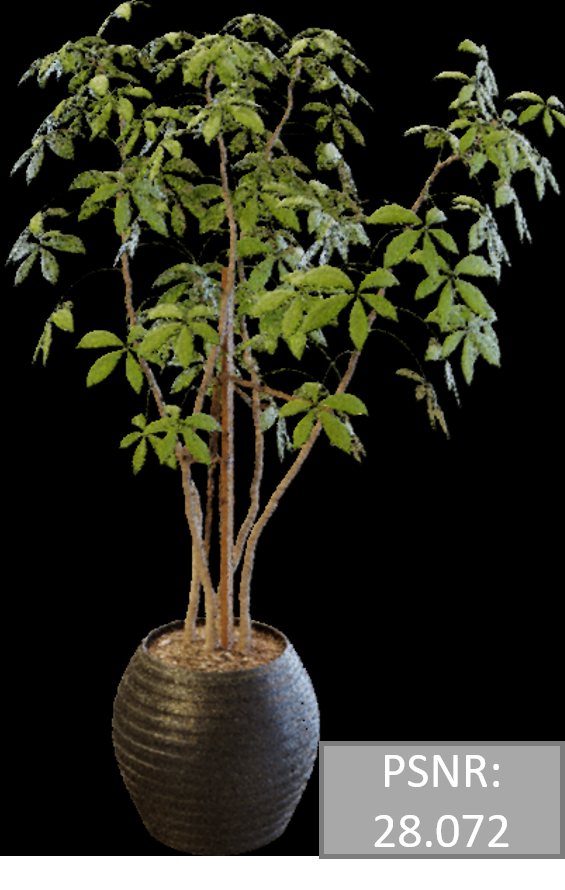}& \includegraphics[align=c,width=0.156\textwidth]{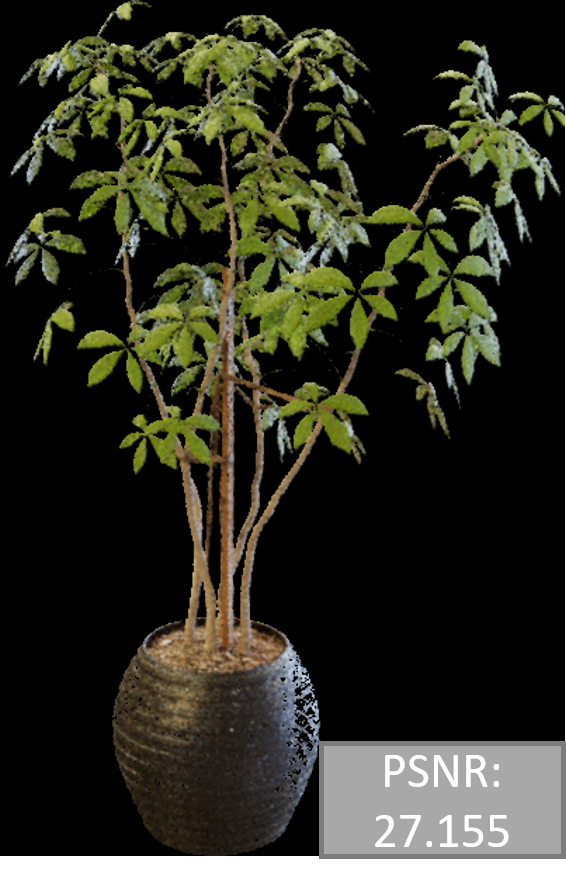} & \includegraphics[align=c,width=0.156\textwidth]{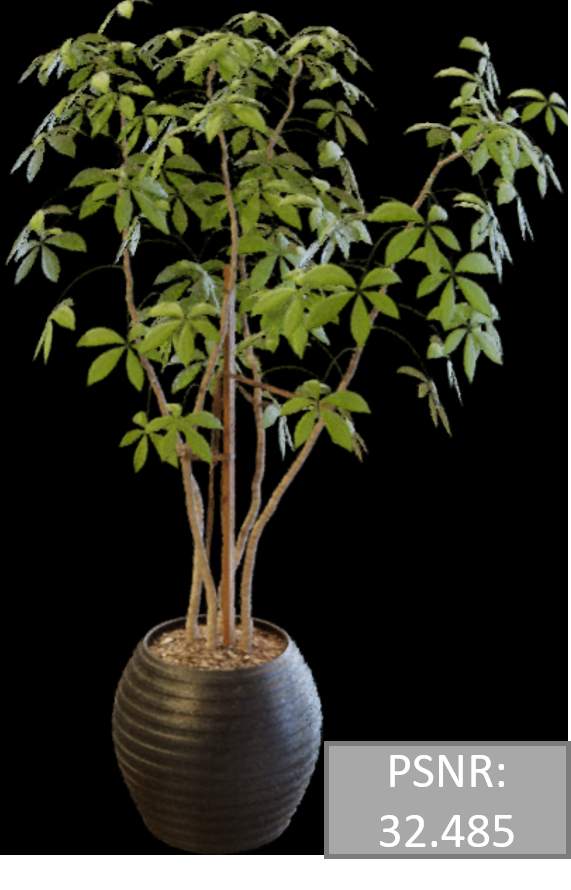} \\

& $10^{-4}$ & $5 \times 10^{-4}$ & $10^{-3}$ & $5 \times 10^{-3}$ & $10^{-2}$ & \textbf{Ours}

\end{tabular}
\caption{\label{fig:comp_threshold}%
    \textbf{Qualitative comparisons on Synthetic-NeRF \cite{mildenhall2020nerf} Ficus data with different thresholds.}
    The corresponding mesh is also visualized in the top row.
    We observe that it is difficult to employ a single maching cubes surface to faithfully represent a NeRF.}

\end{figure*}

\paragraph{Results using Other Teacher Models}
To prove the generalization ability of our method, we also conduct experiments on learning neural duplex radiance fields from other NeRF models.
More specifically, we train Instant-NGP \cite{mueller2022instant} on the NeRF-Synthetic Chair scene \cite{mildenhall2020nerf} from scratch, extract the two geometry proxies using marching cubes with thresholds $-1.5$ and $5.5$ (refer to \cref{fig:ngp_mesh} for geometry visualization).
To ensure the fairness of experiments, we leverage the same sampled camera poses for distillation view synthesis as those used in the main paper, and the same configurations for optimizing neural duplex radiance fields.
We present both quantitative results and qualitative results in \cref{tab: different_teacher} and \cref{fig:teacher_comparison}, which effectively demonstrate the generalizations of the neural duplex radiance fields.

\setlength{\tabcolsep}{13.0pt}
\begin{table*}[t]
\caption{\textbf{Comparisons with different mesh layer settings.} Time: Rasterization overheads of the CUDA version.}
\vspace{-0.8em}
\centering
\scriptsize
\begin{tabular}{@{}r|ccccccc@{}}
Threshold & {A$^{1}$} & {B$^{1}$} & {C$^{1}$} & {\textbf{Ours$^{2}$}} & {D$^{3}$} & {E$^{4}$} & {F}$^{4}$ \\ \hline
$5\times 10^{-5}$ &   &          &              &                &  &   & $\checkmark$       \\
$1\times 10^{-4}$  &    &  $\checkmark$      &             &  $\checkmark$               &  $\checkmark$    & $\checkmark$   &  $\checkmark$    \\
$5 \times 10^{-4}$    &  &          &             &                &        & $\checkmark$ &        \\
$1\times 10^{-3}$ &  $\checkmark$  &       &              &                 & $\checkmark$    &         \\ 
$5 \times 10^{-3}$&    &       &              &                 &     &  $\checkmark$ &       \\ 
$1\times 10^{-2}$ &    &       & $\checkmark$             & $\checkmark$                & $\checkmark$    & $\checkmark$ &   $\checkmark$    \\ 
$5 \times 10^{-2}$ &    &       &             &                &   &   & $\checkmark$        \\ 
\hline

PSNR ${\uparrow}$    & 27.52    &    25.50        &  27.05         &   32.67            &  32.84  &  33.03 &  32.90      \\ 
SSIM ${\uparrow}$    &   0.941  &   0.921        &   0.935         &    0.975            &  0.976 &  0.977  &   0.976      \\ 
LPIPS ${\downarrow}$    &   0.068   &   0.088        & 0.069       &     0.024           & 0.022 & 0.021 &   0.022         \\ 

Time (ms) ${\downarrow}$  &   2.41   &   2.59        & 2.03       &     4.64           &  7.23 & 9.64     &   8.85

\\
\end{tabular}
\label{tab: mesh_layer}
\end{table*}

\begin{figure*}[t!]
\centering
\setlength{\tabcolsep}{1pt}
\begin{tabular}{ccccc}

\includegraphics[align=c,width=0.2\textwidth]{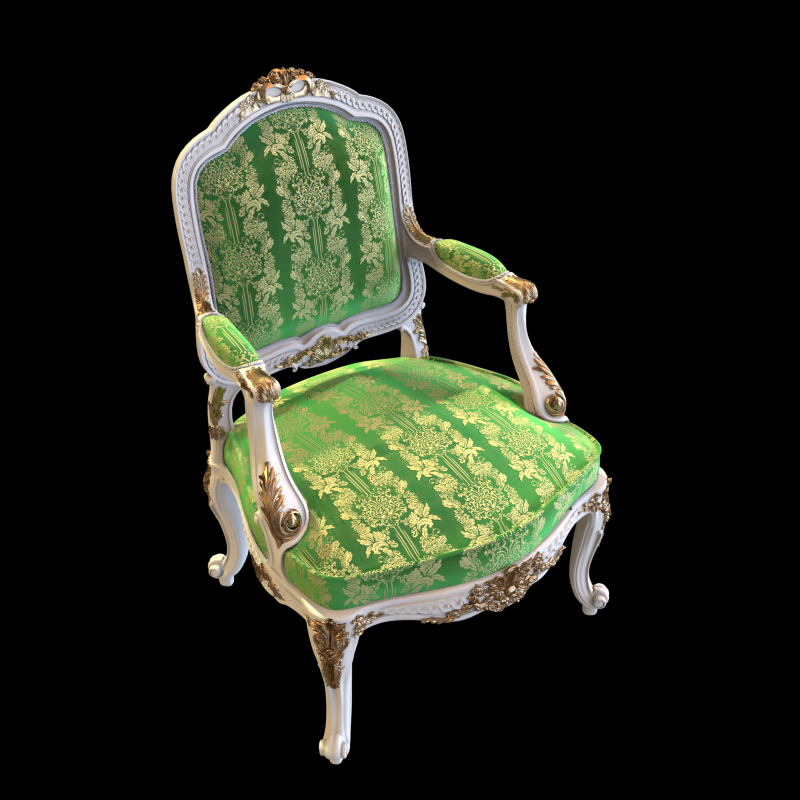} & \includegraphics[align=c,width=0.2\textwidth]{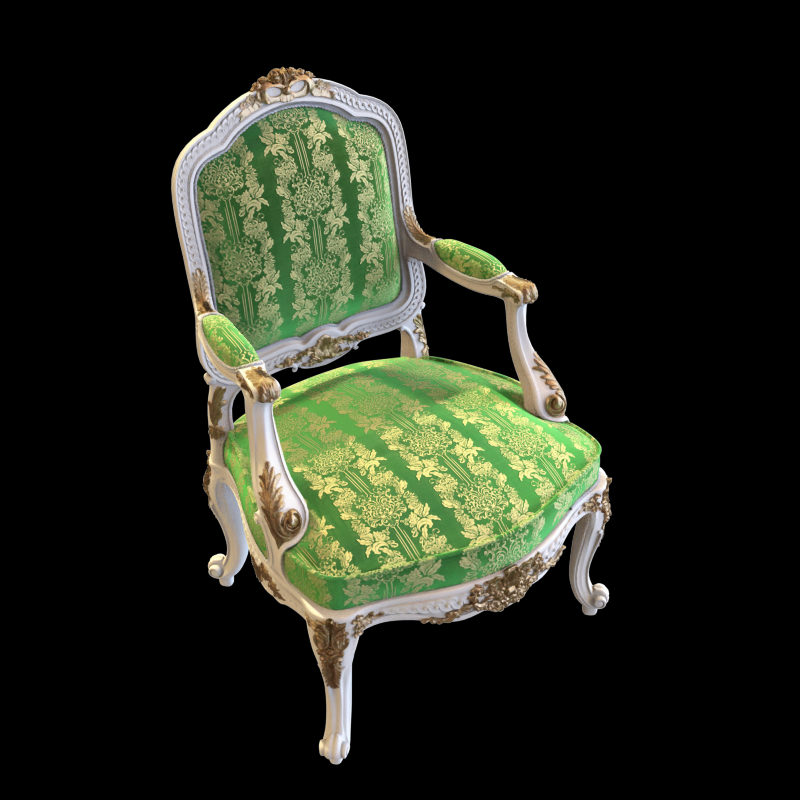} & \includegraphics[align=c,width=0.2\textwidth]{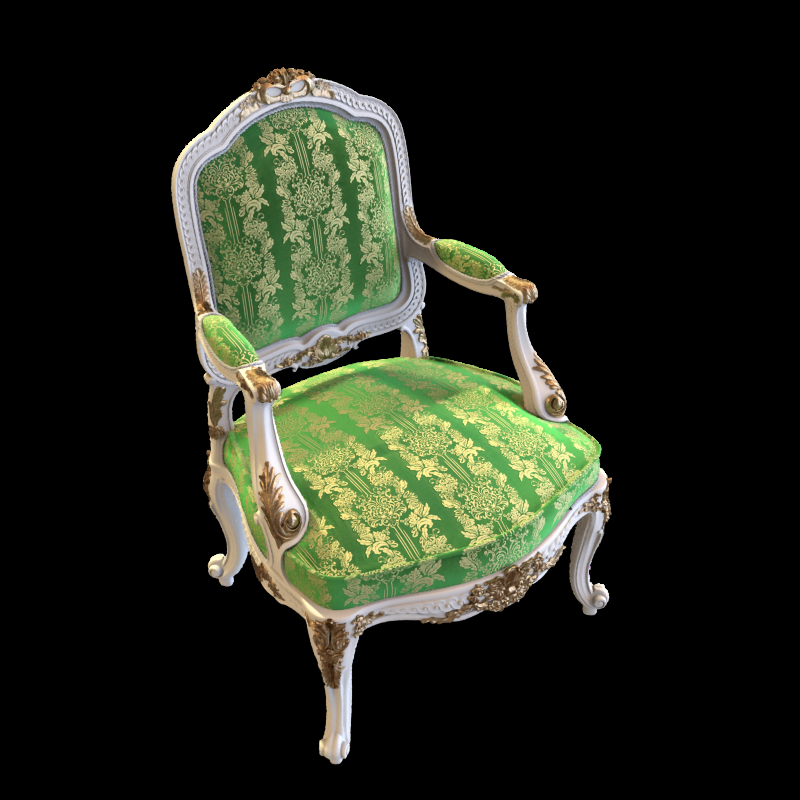} & \includegraphics[align=c,width=0.2\textwidth]{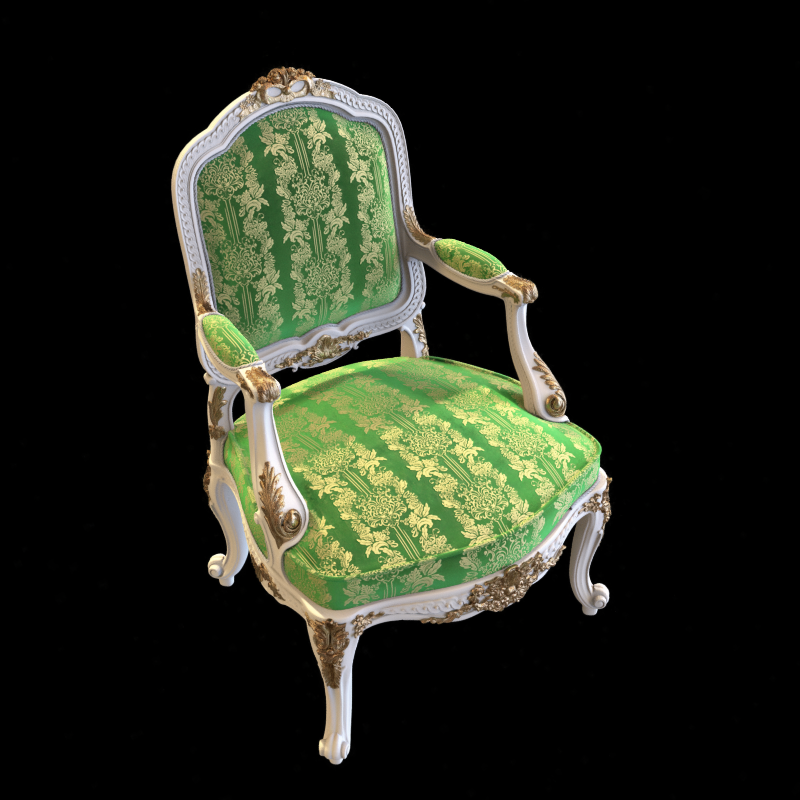} &
\includegraphics[align=c,width=0.2\textwidth]{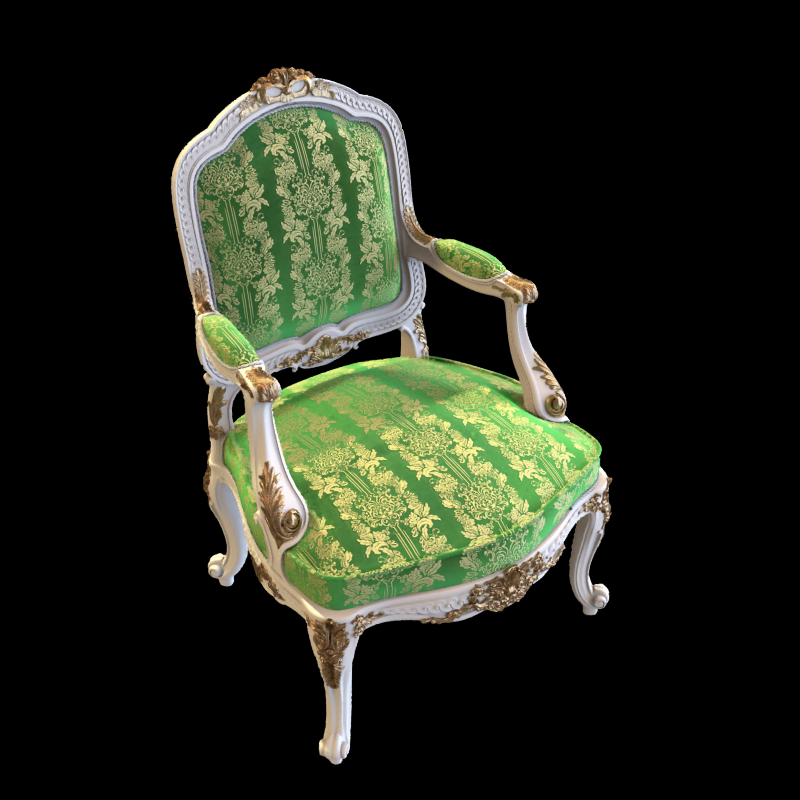}\\

 GT & TensoRF~\cite{Chen2022ECCV} & Ours-TensoRF & Instant-NGP~\cite{mueller2022instant} & Ours-Instant-NGP 
\end{tabular}
\vspace{-2mm}
\caption{\textbf{ Qualitative comparisons on Synthetic-NeRF \cite{mildenhall2020nerf} Chair data using different teacher models.}}
\label{fig:teacher_comparison}
\end{figure*}

\paragraph{Another Perspective to Understand Duplex Radiance Fields}

\begin{figure*}[t!]
\centering
\setlength{\tabcolsep}{1pt}
\begin{tabular}{ccccc}

\includegraphics[align=c,width=0.2\textwidth]{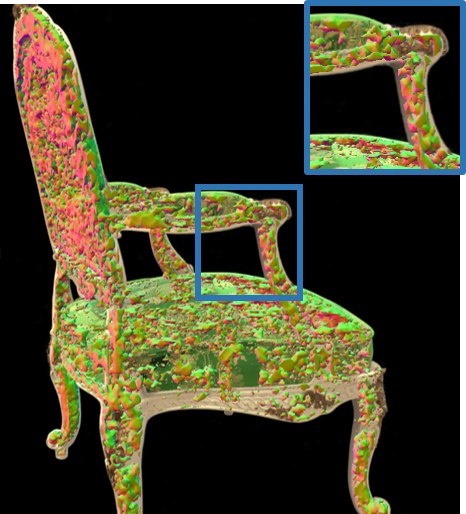} & \includegraphics[align=c,width=0.2\textwidth]{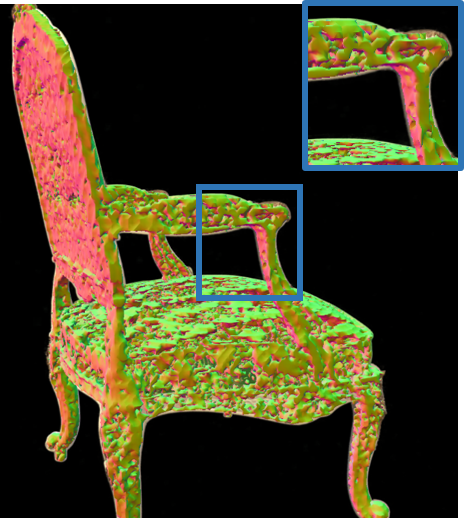} & \includegraphics[align=c,width=0.2\textwidth]{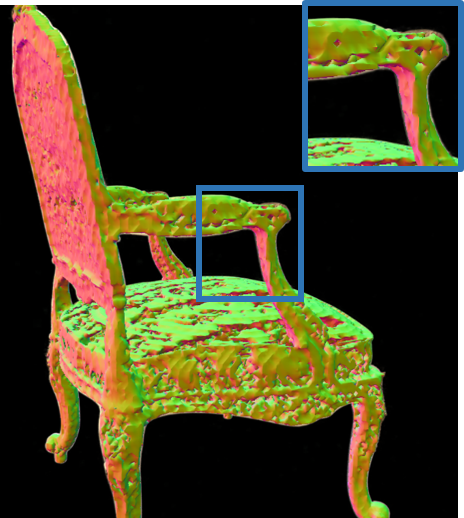} & \includegraphics[align=c,width=0.2\textwidth]{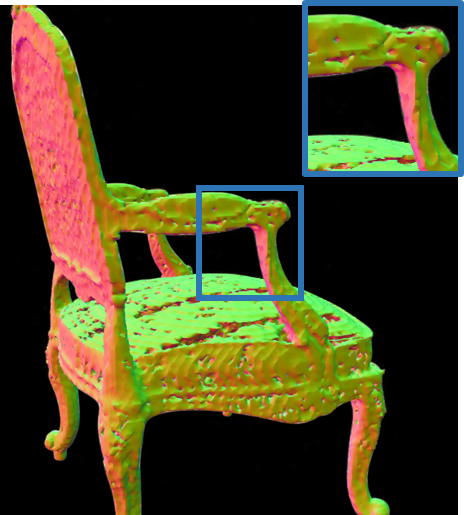} &
\includegraphics[align=c,width=0.2\textwidth]{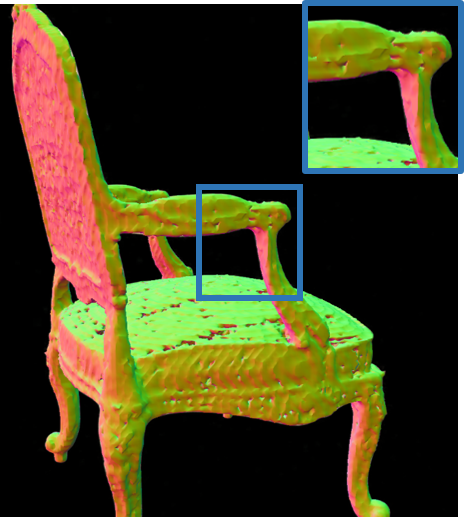}\\

 7.5 & 6.5 & 5.5 & 4.5 & 3.5 \\

\includegraphics[align=c,width=0.2\textwidth]{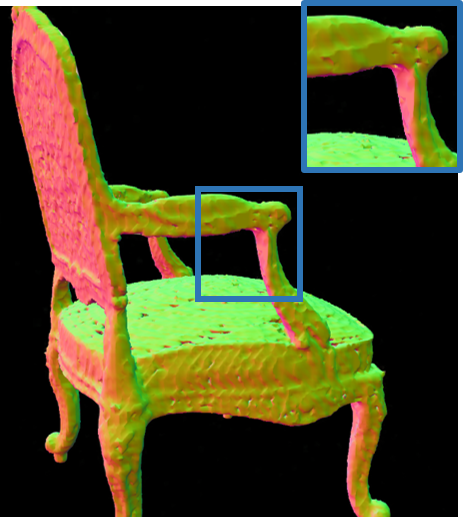} & \includegraphics[align=c,width=0.2\textwidth]{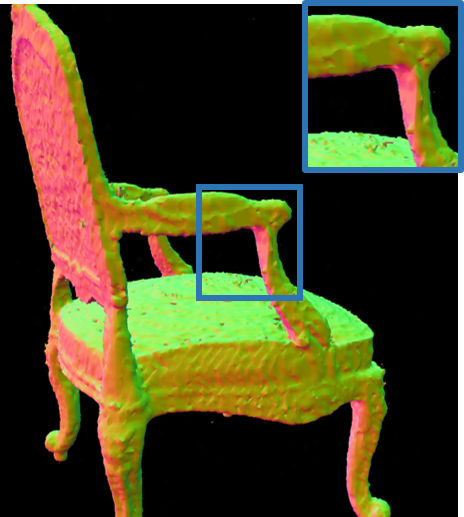} & \includegraphics[align=c,width=0.2\textwidth]{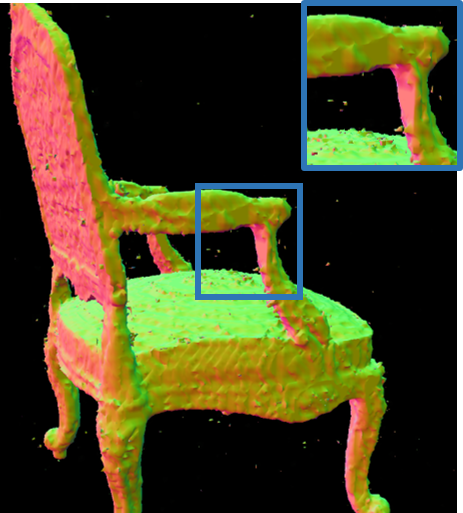} & \includegraphics[align=c,width=0.2\textwidth]{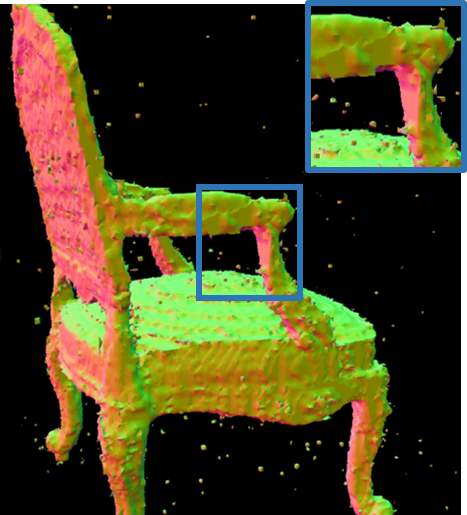}& \includegraphics[align=c,width=0.2\textwidth]{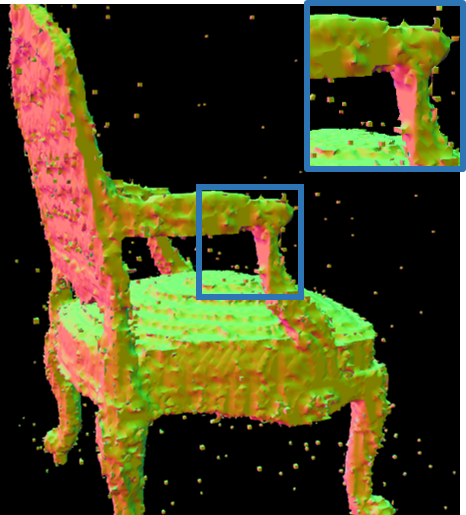}  \\

 2.5 & 1.5 & 0.5 & -0.5 & -1.5 

\end{tabular}
\caption{\textbf{Simultaneous visualization of extracted geometry normal map and corresponding rendering frame. }}
\label{fig:ngp_mesh}
\end{figure*}

To better understand the mechanism of the proposed neural duplex radiance field and how it works, we simultaneously visualize the geometry and the rendering frame under different thresholds from a pretrained Instant-NGP model \cite{mueller2022instant} in \cref{fig:ngp_mesh}.
We have several important observations:
1) Larger thresholds will lead to an under-estimated mesh, which cannot cover the ground-truth surface, but include abundant geometric details.
2) Smaller thresholds will lead to an over-estimated mesh with noise, but it can effectively wrap up the underlying true surface and contains fewer holes.
3) The distribution of the geometries is \emph{spatially heterogeneous}.
For example, the triangle surfaces extracted from threshold in the range $[2.5, 6.5]$ are not concurrently inside or outside of the scenes.
In one mesh, some regions may be covered and others may not.
Hence, based on these observations, it will generally be unlikely that a single surface extracted from a NeRF can represent the 3D scene without sacrificing rendering quality.
In contrast, our proposed neural duplex radiance field can effectively resolve these issues through learning the combination of two-layer duplex meshes while achieving high-quality rendering.

\paragraph{Will more mesh layers lead to better results?}

As shown in \cref{tab: mesh_layer}, increasing the mesh layers can further increase rendering quality, but the gain is \textbf{minor} compared to the improvement from a single mesh (`A'/`B'/`C') to our two-layer duplex mesh (`Ours').
Meanwhile, increasing the mesh layers will result in linear growth of memory footprint and rasterization time, since the hardware rasterization cannot run in parallel for multiple meshes, which will negatively impact the real-time application.
Hence, our duplex radiance field provides the optimal trade-off between quality and speed.

\end{document}

%% file: 1-introduction.tex
\section{Introduction}
\label{sec:intro}

Reconstructing 3D scenes by a representation that can be rendered from unobserved viewpoints using only a few posed images has been a long-standing goal in the computer graphics and computer vision communities.
Significant progress has recently been achieved by neural radiance fields (NeRFs) \cite{mildenhall2020nerf}, which are capable of generating photorealistic novel views and modeling view-dependent effects such as specular reflections.
In particular, a radiance field is a
volumetric function parameterized by MLPs that estimates density and emitted radiance at sampled 3D locations in a given direction.
Differentiable volume rendering then allows the optimization of this function by minimizing the photometric discrepancy between the real observed color and the rendered color.

Despite the unprecedented success and enormous practical potential of NeRF and its various extensions \cite{barron2022mip360, zhang2020nerf++, Chen2022ECCV}, an inescapable problem is the high computational cost of rendering novel views.
For instance, even using a powerful modern GPU, NeRF requires about 30 seconds to render a single image with 800$\times$800\,pixels, which prevents its use for interactive applications in virtual and augmented reality.
On the other hand, the rapid development of NeRF has spawned abundant follow-up works that focus on optimization acceleration \cite{yu2021plenoxels, karnewar2022relu, mueller2022instant}, generalization \cite{yu2021pixelnerf, wang2021ibrnet, chen2021mvsnerf}, and enabled different downstream tasks, including 3D stylization \cite{Huang22StylizedNeRF,FanJWGXW2022,NguyeLX2022}, editing \cite{liu2021editing,yang2021learning,ZhangLWWL2022,yuan2022nerf}, or even perception \cite{jeong2022perfception,DengHYDCYS2022}.
Thus, a generalized method that can learn and extract a real-time renderable representation given an arbitrary pretrained NeRF-based model, while maintaining high rendering quality, is highly desirable.

The huge computational cost of rendering a NeRF representation mainly comes from two aspects:
(1) For each individual pixel, NeRF requires sampling hundreds of locations along the corresponding ray, querying density and then accumulating radiance using volume rendering; and
%
%
(2) A large model size is required to represent the geometric details of complex scenes well, so the MLPs used in NeRF architecture are relatively deep and wide, which incurs significant computation for the evaluation of each point sample.

\setlength{\intextsep}{1pt}
\setlength{\columnsep}{6pt}
\begin{wrapfigure}{r}{0.4\linewidth}
	\includegraphics[width=\linewidth]{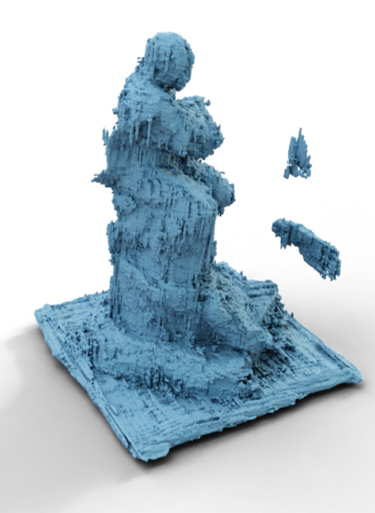}%
 	\vspace{-0.5em}
	\caption{\textbf{NeRF surface.}} 
	\label{fig:noisy_surface}
\end{wrapfigure}

In this paper, we present an approach that achieves high-fidelity real-time view synthesis by addressing these issues. 
%
To avoid the dense sampling along each ray, one solution is to generate a geometry proxy from a pretrained NeRF model, e.g., via marching cubes \cite{lorensen1987marching}.
By exploiting the highly-optimized graphics pipeline, we could almost instantly obtain the sample location for each ray.
However, due to inaccuracies in the density field around surfaces, the extracted mesh may not faithfully represent the true underlying geometry and can contain artifacts, as shown in \cref{fig:noisy_surface}.
Dense local sampling could alleviate these errors to some degree, but cannot handle missing geometries or occlusions.
An alternative approach is to use rasterization for fast neural rendering
\cite{thies2019deferred, riegler2021stable, ruckert2022adop, aliev2020neural} by directly baking neural features onto the surface of explicit the geometry or point cloud.
This is usually accompanied by a deep CNN to translate rasterized features to colors and learn to resolve the existing errors, which is expensive to evaluate and can prevent real-time rendering.

\begin{figure}
\centering
\includegraphics[width=\linewidth]{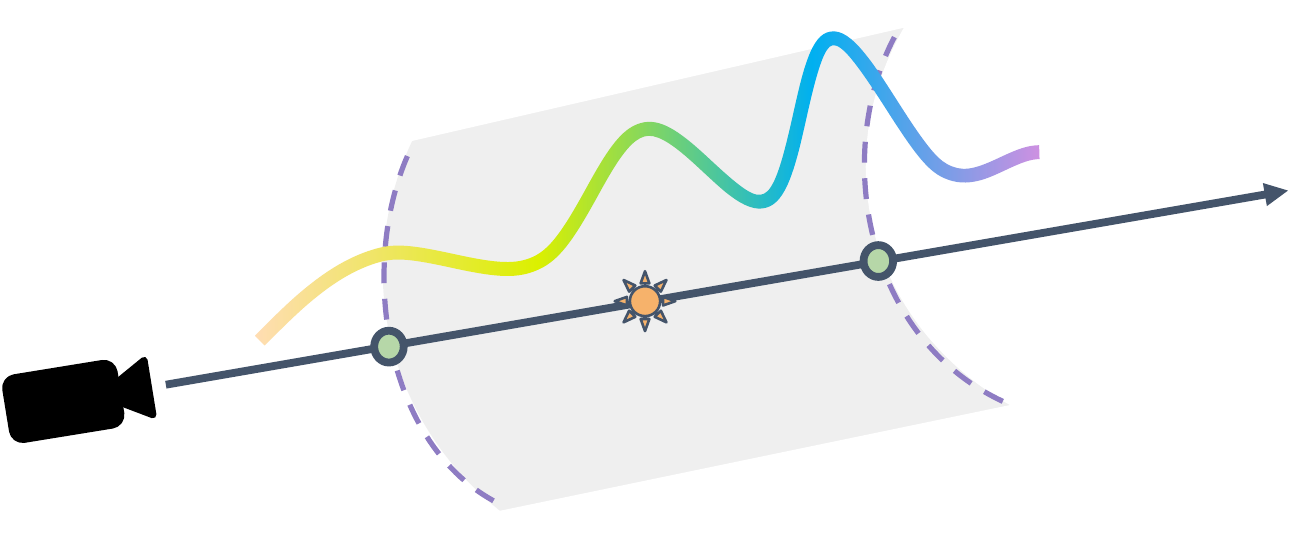}
\caption{\label{fig:sampling}%
    \textbf{Motivation.} 
    The continuous density distribution of NeRF (curve above the ray) makes it difficult to identify the accurate location of a surface (\raisebox{-2pt}{\textcolor{Apricot}{\EightStar}}) for appearance calculation.
    We thus seek to extract a reliable interval from the density field (between dashed lines), and learn the duplex radiance combinations to tolerate errors.
}
\vspace{-0.5em}
\end{figure}

To significantly reduce the sampled numbers while efficiently handling the geometric errors,
our \textit{first} key technical innovation is to infer the final RGB color according to the radiance of duplex points along the ray.
As illustrated in \cref{fig:sampling}, NeRFs represent a continuous density distribution along each ray.
%
While it will generally be difficult to determine the exact location of a surface along the ray, it will be easier to extract a reliable interval that contributes the most to the final prediction by using an under- and an overestimation of the geometry.
%
Motivated by this idea, instead of selecting a specific location for appearance calculation, or performing expensive dense volume rendering in this interval, we represent the scene using learnable features at the two intersection points of a ray with the duplex geometry and use a neural network to learn the color from this aggregated duplex radiance information.
Although only two sampled locations are considered, we found this proposed neural duplex radiance field to be robust in compensating for the errors of the geometry proxy even without a deep neural network, while effectively preserving the efficiency of rasterization-based approaches.
The NeRF MLP has become the most standard architecture for most neural implicit representations \cite{chen2019learning,sitzmann2019scene,oechsle2019texture,mildenhall2020nerf,mildenhall2021nerf,chen2021learning}. Yet, with only a few points considered along the ray, the MLP struggles to constrain the proposed neural duplex radiance field.
\textit{Instead}, we use a shallow convolutional network, which can effectively capture the local geometric information of neighboring pixels, and leads to a considerably better rendering quality.
\textit{Finally}, we found that directly training the neural duplex radiance field from scratch will lead to noticeable artifacts.
We therefore propose a multi-view distillation optimization strategy that enables us to effectively approximate the rendering quality of the original NeRF models.
Remarkably, our method improves run-time performance by \emph{10,000} times compared to the original NeRF while maintaining high-quality rendering.

%% file: 2-relatedwork.tex
\begin{figure*}
\centering
\includegraphics[width=\linewidth]{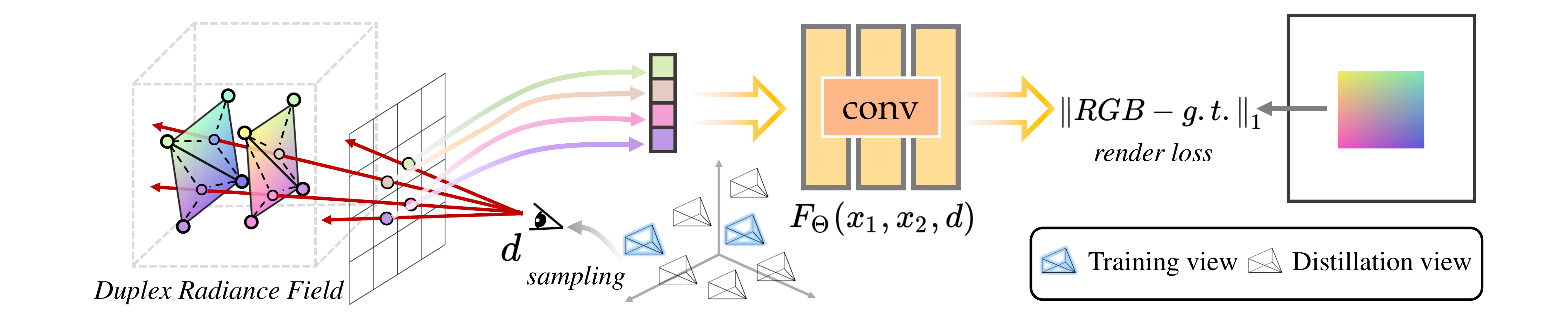}
\caption{\label{sec:pipeline}%
    \textbf{Architecture of Neural Duplex Radiance Fields.}
    Our neural representation builds on a two-layer duplex surface extracted from a NeRF model, which approximates the reliable interval covering the true underlying geometry.
    For each sampled target camera ray with view direction $d$, we find the duplex ray-surface intersections and derive the radiance features using barycentric interpolation.
    We use a shallow CNN to aggregate the local geometry and appearance information, and produce view-dependent colors, which effectively ensures rendering efficiency.
    The output color is directly compared with the pixels of rendered or ground-truth views to optimize the whole representation.
}
\vspace{-0.5em}
\end{figure*}

\section{Related Work}
\label{sec:related_works}

Our work is in the category of neural rendering.
Since the introduction of the seminal NeRF approach \cite{mildenhall2020nerf}, the growth of neural rendering papers is exponential \cite{tewari2022advances}.
Here, we focus on works closely related to our method along two directions: \textit{neural representations} and \textit{real-time neural rendering}.

\paragraph{Neural Representations}

NeRF \cite{mildenhall2020nerf} represents the scene as a continuous field of density and color, and generates images by volume rendering.
Built upon NeRF, Mip-NeRF \cite{barron2022mip360} represents the prefiltered radiance field for a continuous space of scales, which enables accuracy improvements for different resolutions.
NeRF++ \cite{zhang2020nerf++} and Mip-NeRF 360 \cite{barron2022mip360} further extend NeRF-like models for unbounded scenes, allowing reconstruction of scenes with far-away backgrounds.
NeuS \cite{wang2021neus} and VolSDF \cite{yariv2021volume} propose to learn a signed distance function representation by volume rendering to better reconstruct 3D objects.
This has potential for fast surface rendering, but the visual quality may not reach the photorealism of NeRF.
PixelNeRF \cite{yu2021pixelnerf}, IBRNet \cite{wang2021ibrnet} and MVSNeRF \cite{chen2021mvsnerf} learn more generalized neural representations to overcome NeRF's inability to share knowledge between scenes and can achieve high-quality novel-view synthesis given sparse input images.
More recently, efforts are made to extremely reduce the training time of NeRF such as Instant-NGP \cite{mueller2022instant}, using multiresolution hash encoding for training acceleration, or TensoRF \cite{Chen2022ECCV}, using tensor factorization to enable memory-efficient feature storage and so on \cite{yu2021plenoxels, SunSC2022, karnewar2022relu}. Neural light field networks \cite{SitzmRFTD2021} learn to directly map rays to observed radiance, and thus only require a single evaluation per pixel. Recently, multiple ways have been proposed to parameterize camera rays, such as Plücker coordinates \cite{SitzmRFTD2021}, Gegenbauer polynomials \cite{feng2021signet}, or local affine embeddings \cite{attal2022learning}.
Albeit with great progress, neural light fields still struggle to achieve view-consistent and high-quality 360-degree rendering due to the lack of an explicit geometry prior.

\paragraph{Real-Time Neural Rendering}


There are various directions to accelerate NeRF rendering by reducing the considerable computation needed for inference of large networks and dense sampling along view rays.
For example, by deriving sparse intermediate representations from NeRF, such as sparse octrees \cite{yu2021plenoctrees, wang2022fourier} or sparse voxel grids \cite{hedman2021snerg}, by dividing the space into many smaller MLPs \cite{reiser2021kilonerf, rebain2021derf}, or by adopting an efficienct sampling strategy
\cite{neff2021donerf, hu2022efficientnerf, KurzNLZS2022}.
Inspired by the rendering equation in computer graphics, FastNeRF \cite{garbin2021fastnerf} proposes to factorize the NeRF and uses caching to achieve efficient rendering.
These voxel-based methods basically leverage the caching concept to skip abundant computations in exchange for a larger memory footprint. 
NeX \cite{wizadwongsa2021nex} proposes to combine the MPI representation and neural basis functions to enable real-time view synthesis for forward-facing data.
EG3D \cite{chan2022efficient} targets 3D-aware synthesis and supports real-time application by low-resolution rendering followed by upsampling. ENeRF \cite{lin2022efficient} tries to tackle interactive free-viewpoint video by skipping the sampling of empty space. More recently, concurrent work MobileNeRF \cite{ChenFHT2023} utilizes optimized sheet meshes as proxy geometry to achieve highly efficient rendering on commodity graphics hardware.
In contrast to MobileNeRF \cite{ChenFHT2023}, our approach neither requires complex multi-stage optimization nor retraining the scene from scratch; instead, our approach can serve as a post-processing step for any existing NeRFs.
%
Another direction explores alternative representations instead of functional NeRFs, such as point clouds \cite{aliev2020neural, rakhimov2022text, ruckert2022adop} or spheres \cite{lassner2021pulsar}.

%% file: 3-method.tex
\section{Method}
\label{sec:method}

Our main goal is to learn an efficient 3D scene representation that enables high-quality, real-time novel-view synthesis based on a given NeRF model.
To this end, we introduce the neural duplex radiance field (NDRF), a novel concept to significantly decrease the number of sampled points along a ray from hundreds to two, while preserving the realistic details of complex scenes well (\cref{sec:duplex}).
Although the query frequency is bounded by using a two-layered mesh, the per-sample deep MLP-based forward pass in NeRF is still burdensome and lacks expressivity, especially for high-resolution rendering scenarios.
Thus, we propose a more efficient hybrid radiance representation and shading mechanism (\cref{sec:conv-shading}).
To further suppress potential artifacts and improve the rendering quality, we leverage multi-view distillation information to guide the optimization procedure (\cref{sec:multiview-distillation}).
Finally, we demonstrate the complete rendering procedure and our shader-based implementation that enables real-time cross-platform rendering (\cref{sec:rendering}). 

\subsection{Neural Duplex Radiance Fields}
\label{sec:duplex}


Let us start with a quick review of NeRF basics.
In a nutshell, NeRF approximates the 5D plenoptic function with a learnable MLP, which maps a spatial location $\mathbf{x} \!\in\! \mathbb{R}^3$ and view direction $\mathbf{d} \!\in\! \mathbb{S}^2$ to radiance $\mathbf{c} \!\in\! \mathbb{R}^3$ and density $\sigma \!\in\! \mathbb{R}$.
After casting a ray from the camera center through a pixel into the scene, its color is calculated by evaluating a network $F_\Theta$ at multiple locations $\mathbf{x}$ along the ray, and aggregating the radiance using volume rendering \cite{mildenhall2020nerf}.

\begin{figure}
\centering
\includegraphics[width=\linewidth]{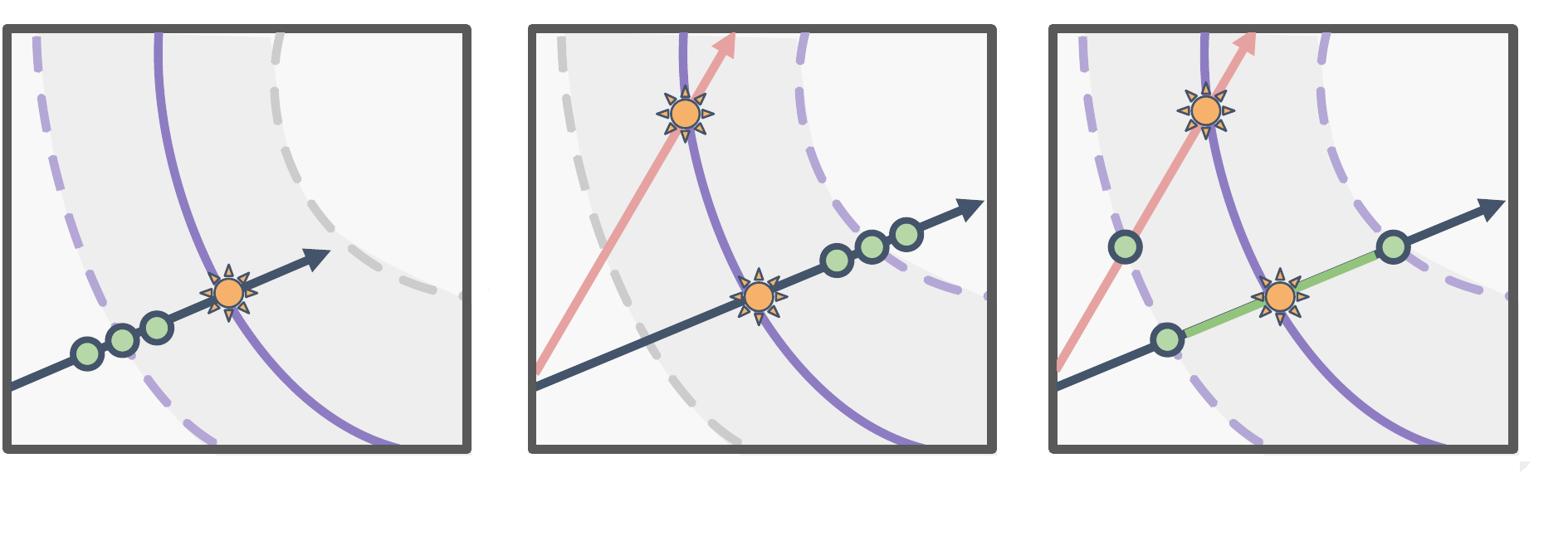}
\vspace{-1.8em}
\caption{\label{fig:why_need_duplex}%
    \textbf{Duplex radiance field vs local sampling.}
    \raisebox{-2pt}{\textcolor{Apricot}{\EightStar}}: real ray-mesh intersection.
    \textcolor{YellowGreen}{$\bullet$}: local sampled points around proxy surface.
    Dashed curve: estimated surface.
    \textcolor{Periwinkle}{Solid curve}: ground-truth surface.
    \textcolor{gray}{Gray curve}: not currently estimated surface.
    \textbf{Left:} The estimated surface occludes the appearance location on the actual surface.
    \textbf{Middle:} The \textcolor{Salmon}{red ray} may not hit the estimated surface, resulting in hole artifacts or incorrect intersections.
    \textbf{Right:} Our duplex radiance field instead learns to combine locations on two estimated surfaces to render the correct color, and effectively decreases ray misses.
    %
    %
    %
    %
    }
\vspace{-1.0em}
\end{figure}

A key factor that impacts the rendering efficiency of NeRF is the number of sampled 3D locations for each ray.
Since the volumetric representation of NeRF can be converted into a triangle mesh via marching cubes, a natural idea is to directly bake the appearance on a geometric surface, \ie, to only consider the radiance of ray-surface intersections.
However, due to the discretization error of marching cubes and the inherent floater artifacts of NeRF \cite{barron2022mip360},
the generated mesh quality is often inadequate (see \cref{fig:noisy_surface}).
Rasterization-based neural rendering methods overcome imperfections in the 3D proxy by correcting errors in screen space using a deep CNN, which inevitably sacrifices rendering efficiency \cite{Riegler2020FVS, riegler2021stable}.
%
Since the geometric proxy already conveys the approximate 3D structure, an alternative way would be locally sampling around the surface.
Although this can help counter the low-quality mesh, it struggles with missing geometry or visible occlusions, as illustrated in \cref{fig:why_need_duplex}.

\paragraph{Representation}

It is worth noting that the density distribution of NeRF, which determines the contribution of emitted radiance to the final color of a ray, may contain multiple peaks.
%
This makes approximating the color of a pixel using a single ray--surface intersection a challenging proposition.
To significantly reduce the sampled points for each pixel and simultaneously ensure the rendering fidelity, we propose the neural duplex radiance field $F_\Theta: \mathbb{R}^8 \mapsto \mathbb{R}^3$ to represent a light ray, where $F_\Theta$ learns to map a ray segment defined by its two endpoints, $\mathbf{x}_1$ and $\mathbf{x}_2$, to integrated view-dependent colors.
%
More specifically, our goal is to directly learn the color of any pixel from the ray segment defined by the reliable interval of the density distribution.
We acquire the endpoints of a ray segment by casting the ray into two proxy mesh surfaces, the inner $\mathcal{M}_\text{i}(\mathcal{V}_\text{i}, \mathcal{T}_\text{i})$ and outer $\mathcal{M}_\text{o}(\mathcal{V}_\text{o}, \mathcal{T}_\text{o})$, with vertex positions $\mathcal{V}_\bullet$ and triangle faces $\mathcal{T}_\bullet$.
These meshes are extracted from different level sets of NeRF's density field.
More formally, the ray color $\mathbf{c}$ could be calculated by
\begin{align}
    \mathbf{c} = F_{\Theta}(\mathbf{x}_1, \mathbf{x}_2, \mathbf{d}) \quad \text{for} \quad \mathbf{x}_1 \in \mathcal{T}_o \text{ and } \mathbf{x}_2 \in \mathcal{T}_i \text{,}
\end{align}
where $\mathbf{x}_1$ and $\mathbf{x}_2$ are the depth-sorted intersections of a given ray with meshes $\mathcal{M}_\text{i}$ and $\mathcal{M}_\text{o}$, and $\mathbf{d}$ denotes the view direction.
We still append the view direction considering not all rays will hit both surface proxies.
The explicit triangle mesh then will allow us to efficiently acquire the ray–surface intersection, benefiting from the full parallelism of the graphics rasterization pipeline.

In most implicit neural representations, different properties like SDF, radiance, or pixel colors are encoded by a multilayer perceptron due to its smoothness and compactness properties.
However, densely evaluating a deep MLP is not a wise choice for real-time applications considering the high computational cost of inference.
Since our radiance field is defined via two geometric surfaces, our encoding is defined by directly attaching learnable features $\mathbf{f}_k \!\in\! \mathbb{R}^N$ to each mesh vertex $\mathbf{v}_k$.
To obtain features $\mathbf{f}(\mathbf{x})$ for any point $\mathbf{x}$ on a triangle, we interpolate features within triangles using barycentric coordinates via hardware rasterization.
Thus, the rendering equation can be re-written as
\begin{align}
    \mathbf{c} = F_{\Theta}(\mathbf{f}(\mathbf{x}_1), \mathbf{f}(\mathbf{x}_2), \mathbf{d})  \quad \text{with} \quad \mathbf{x}_1 \in \mathcal{T}_o \text{, } \mathbf{x}_2 \in \mathcal{T}_i \text{,}
\end{align}
%
where $\Theta$ parameterizes a shallow network to learn the aggregation of radiance features given each ray.


\subsection{Convolutional Shading}
\label{sec:conv-shading}

Many NeRF-based methods use a view direction conditioned MLP to learn view-dependent appearance effects.
%
The dense sampling along rays provides a feasible way to constrain the learning of the radiance fields through shared 3D locations, which ultimately enables high-quality and coherent rendering across different views. This implicit constraint, however,  does not apply to our case.
To highly optimize run-time efficiency, neural duplex radiance only considers two points for each cast ray, which significantly reduces the capability of capturing internal correlation in training, often resulting in spatial artifacts and the degradation of rendering quality.
%

We resolve this issue by a convolutional shading network that converts duplex features and view directions to per-pixel colors in image space using a small receptive field.
More specifically, an independent pixel will be rendered by considering the neighboring local geometry positions $\mathbf{x}_i$ whose projection $\mathbf{K}\left(\mathbf{R} \mathbf{x}_i+\mathbf{t}\right)$ with the nearest $z$-coordinate lies in the local window around this pixel location, where camera pose $\mathbf{R} \!\in\! \mathrm{SO}(3)$, $\mathbf{t} \!\in\! \mathbb{R}^3$ and $\mathbf{K}$ is the intrinsic matrix. This approach effectively increases the correlation of samples of a 3D scene.
To ensure high inference speed, we only use 2$\sim $3 convolutional layers and limit convolution kernels to a small 2$\times$2 window for optimal visual fidelity and view consistency.
Please refer to the supplemental material for more detailed network architecture and parameters.

\subsection{Multi-View Distillation}
\label{sec:multiview-distillation}

We observed that neural duplex radiance fields trained from scratch fail to match the high fidelity of the original NeRFs and produce many spatial high-frequency artifacts.
To overcome this problem, we take inspiration from KiloNeRF \cite{reiser2021kilonerf} and use a pretrained NeRF as a teacher model to guide the optimization of our neural duplex radiance fields.
%
KiloNeRF directly minimizes the differences of density and radiance in 3D volume space between teacher and student models without any rendering.
Considering different parameterizations of NeRF and our models, we instead use rendered multi-view images to distill knowledge from the teacher model.

One important question is how to generate effective distillation views for rendering based on the training dataset.
For object-level datasets, camera views tend to lie on a hemisphere that can be estimated and used for sampling new distillation viewpoints similar with R2L \cite{wang2022r2l}.
For real-world captures, the camera poses may only occupy a small subspace.
To produce meaningful interpolated viewpoints under both settings, we first align the view directions of all cameras with the origin and convert the camera poses into Spherical coordinates $(r, \theta, \varphi)$ relative to the origin.
For each distillation view, we then randomly sample a radius $r_s \!\sim\! U\!\left(r^{\min}, r^{\max}\right)$, angles $\theta_s \!\sim\! U\!\left(\theta^{\min}, \theta^{\max}\right)$ and $\varphi_s \!\sim\! U\!\left(\varphi^{\min}, \varphi^{\max}\right)$ according to the ranges of spherical coordinates.
%
%
%
After converting from spherical to Cartesian coordinates, we obtain many suitable interpolated poses following the original viewpoint distribution. We define the number of distillation views to be 1,000 in all settings.
%
Our model will be trained on the distilled views first and then fine-tuned using the training images to produce sharper appearance.

\begin{figure}[!t]
\centering
\includegraphics[width=\linewidth]{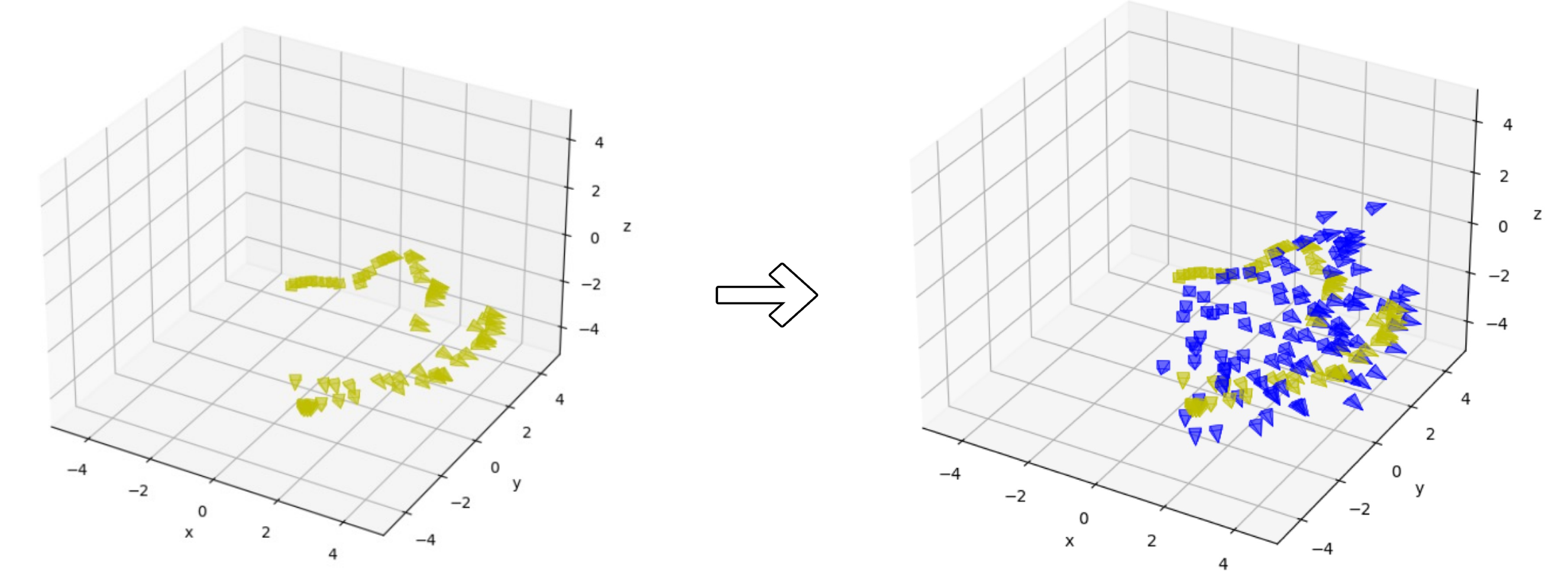}
\vspace{-1.8em}
\caption{\label{fig:pose_generation}%
\textbf{Visualizations of distillation view generation.}
\textcolor{GreenYellow}{Yellow cameras}: training views.
\textcolor{blue}{Blue cameras}: sampled distillation views.}
\vspace{-1em}
\end{figure}

\subsection{Real-Time Rendering}
\label{sec:rendering}


A fully trained neural duplex radiance field comprises two meshes with per-vertex features as well as the shallow shading CNN.
%
While we train the model using PyTorch \cite{PaszkGMLBCKLGADKYDRTCSFBC2019}, we implemented a real-time renderer in WebGL using GLSL shaders for cross-platform compatibility.
%
To synthesize a novel view, we use hardware rasterization to efficiently render two feature buffers at the output image resolution.
These buffers are sent into the CNN together with a per-pixel view direction to aggregate the local radiance features and produce view-dependent RGB colors.
%
For efficiency, we implement the CNN in two rendering passes, one pass for each layer.
%
Thanks to the compact size of the convolution kernels, each convolution layer can be executed in a pixel-parallel way, which is highly efficient, even without CUDA.

%% file: 4-experiments.tex
\section{Experiments}

\subsection{Setup}

\paragraph{Datasets}

We evaluate and compare performance on the NeRF-Synthetic \cite{mildenhall2020nerf} benchmark dataset, which contains eight objects with 360° views at 800$\times$800 resolution.
Further, to comprehensively compare the quality and to test the upper bound of speed, we conduct experiments on the megapixel-level dataset of Tanks\&Temples \cite{knapitsch2017tanks}, which contains 1920$\times$1080 frames.
%
%
We also leverage additional real-world datasets to perform ablation studies.
All the training and testing splits follow the published literature.

\paragraph{Baselines}

We compare with two representative efficient novel-view synthesis approaches: KiloNeRF \cite{reiser2021kilonerf} and SNeRG \cite{hedman2021snerg}.
For quantitative comparisons, we directly use the published numerical results.
For qualitative comparisons, we retrain both methods on different datasets following original configurations, since not all checkpoints are released.

\begin{table*}
\caption{\label{tab:quantitative_1}%
   \textbf{ Quantitative results on Synthetic-NeRF \cite{mildenhall2020nerf} and Tanks\&Temples \cite{knapitsch2017tanks} datasets.}
    The best result is highlighted in \textbf{bold}.
    Ours (\textit{laptop}): Run on an M1 MacBook Pro with 16\,GB RAM. Others run on a desktop. N+M: N is the training time and M is the baking time.
}
\vspace{-0.5em}
\centering
\setlength{\tabcolsep}{2.8mm}
\begin{tabular}{llcrccccc}
\toprule
Dataset                                                                              & Method   & Backend & FPS $\uparrow$    & PSNR $\uparrow$ & SSIM $\uparrow$ & LPIPS $\downarrow$ & \begin{tabular}[c]{@{}c@{}}Training\\ (hours) \end{tabular} & \begin{tabular}[c]{@{}c@{}}Training\\ Specs \end{tabular}\\ \midrule
\multirow{5}{*}{\begin{tabular}[c]{@{}l@{}}Synthetic-NeRF\\ 800$\times$800\end{tabular}}
& KiloNeRF~\cite{reiser2021kilonerf}  &  CUDA   &      33.47  &     31.00  &     0.950  & \bf 0.030  & $\sim$35+14 & 1$\times$V100  \\
& SNeRG~\cite{hedman2021snerg}        &  WebGL  &      89.43  &     30.38  &     0.950  &     0.050  & $\sim$30+5  & 8$\times$A100  \\ 
\cmidrule{2-9}
& Ours                                &  CUDA   &      59.71  & \bf 32.14  & \bf 0.957  & \bf 0.030  &     6.61    & 1$\times$3090  \\
& Ours                                &  WebGL  & \bf 282.67  &     30.43  &     0.945  &     0.049  & \bf 1.01    & 1$\times$3090  \\
& Ours (\textit{laptop})              &  WebGL  &      62.27  &     30.43  &     0.945  &     0.049  & \bf 1.01    & 1$\times$3090  \\
\midrule
\multirow{5}{*}{\begin{tabular}[c]{@{}l@{}}Tanks\&Temples\\ 1920$\times$1080\end{tabular}}
& KiloNeRF  & CUDA    &      16.30  & \bf 28.41  &     0.910  &     0.090  & $\sim$24+24 & 1$\times$A100  \\
& SNeRG        & WebGL   &      55.48  &     24.95  &     0.882  &     0.184  & $\sim$30+5  & 8$\times$A100  \\
\cmidrule{2-9}
& Ours                                & CUDA    &      27.49  &     28.12  & \bf 0.915  & \bf 0.089  &     9.02    & 1$\times$A100  \\
& Ours                                & WebGL   & \bf 186.94  &     27.08  &     0.895  &     0.130  & \bf 1.84    & 1$\times$A100  \\
& Ours (\textit{laptop})                      & WebGL   &      33.64  &     27.08  &     0.895  &     0.130  & \bf 1.84    & 1$\times$A100  \\
\bottomrule
\end{tabular}
\vspace{-1em}
\end{table*}

\paragraph{Evaluation Metrics}

We measure rendering quality by comparing predicted novel views and ground-truth images using standard metrics:
peak signal-to-noise ratio (PSNR), structural similarity index (SSIM) \cite{wang2004image}, and learned perceptual image patch similarity (LPIPS) \cite{zhang2018unreasonable}.
Since our goal is highly efficient rendering, we also report framerates using frames per second (FPS).
%
We also compare the GPU memory consumption in GB for WebGL applications, to quantify the trade-off between speed, quality, and memory.
Finally, we compare the training speed by recording the total optimization time in hours.
Our test system consists of an NVIDIA RTX 2080 Ti GPU, an i7-9700K CPU, and 32 GB RAM.

\paragraph{Implementation}

We implement the CUDA version with 20-dimensional learnable radiance features and a three-layer convolutional network.
To ensure high efficiency and match the shader implementation, in the WebGL version, we decrease the learnable feature dimension attached to the geometry surface from 20 to 8 and remove the final convolution layer.
This results in some performance degradation on both datasets compared with our CUDA version, but brings remarkable FPS improvement and cross-platform properties, since the renderer implementation no longer requires CUDA.
More details regarding the architecture design and parameters setting can be found in the supplementary material.

\paragraph{Optimization Details}

We build our method on top of PyTorch \cite{PaszkGMLBCKLGADKYDRTCSFBC2019} and use the Adam optimizer \cite{kingma2014adam} with $\beta_1=0.9$ and $\beta_2=0.999$.
Our method converges in 200,000 iterations, where the first half is optimized on the distillation views and the second half on the original training images.
In each iteration, we randomly sample one view, cast all rays into the geometry to fetch the duplex radiance features, and generate complete screen-space buffers (or local patches if training memory is insufficient) for the convolutional shading.
The learning rate is set to $10^{-3}$ initially, with exponential decay to $10^{-5}$ thereafter.
%
We chose TensoRF \cite{Chen2022ECCV} as the teacher NeRF model considering its high efficiency of reconstruction and synthesis of distillation views.
%
To convert the density field into the duplex mesh geometry, the two thresholds are set empirically to $10^{-4}$ and $10^{-2}$, respectively.

\begin{table}
\caption{\label{tab:mem_webgl}%
    \textbf{Memory--Quality--Speed trade-off.}
    Peak GPU memory consumption (GB), rendering quality, run-time performance comparisons on Tanks\&Temples~\cite{knapitsch2017tanks} dataset for WebGL applications.
}
\vspace{-0.5em}
\centering
\setlength{\tabcolsep}{3.0mm}
\resizebox{\linewidth}{!}{%
\begin{tabular}{lrrr}
\toprule
Method       & Memory (GB) ${\downarrow}$   & PSNR  ${\uparrow}$         & FPS  ${\uparrow}$ \\
\midrule
SNeRG        & 4.19          & 24.95          & 55.48           \\
Ours         & 0.99          & \textbf{27.08} & \textbf{186.94} \\
Ours (\textit{laptop}) & \textbf{0.94} & \textbf{27.08} & 33.64 \\
\bottomrule
\end{tabular}%
}
\end{table}

\subsection{Results}

\paragraph{Quantitative Comparison}

\begin{figure*}[tbp]
\begin{center}
\setlength{\tabcolsep}{1pt}
\begin{tabular}{ccccc}

\rotatebox[origin=c]{90}{800 $\times$ 800 \strut} & 
\includegraphics[align=c,width=0.235\textwidth]{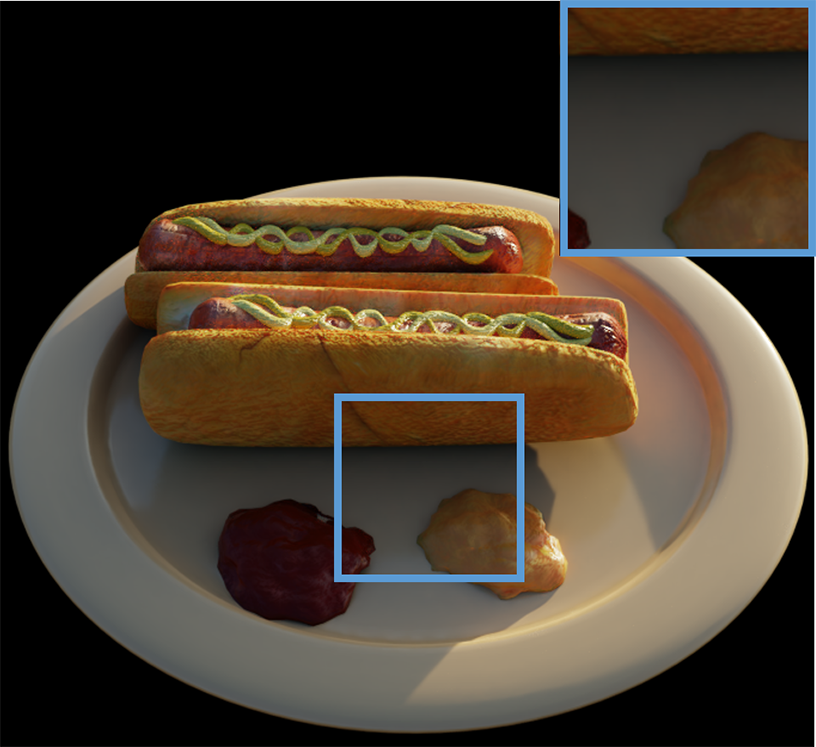} & \includegraphics[align=c,width=0.235\textwidth]{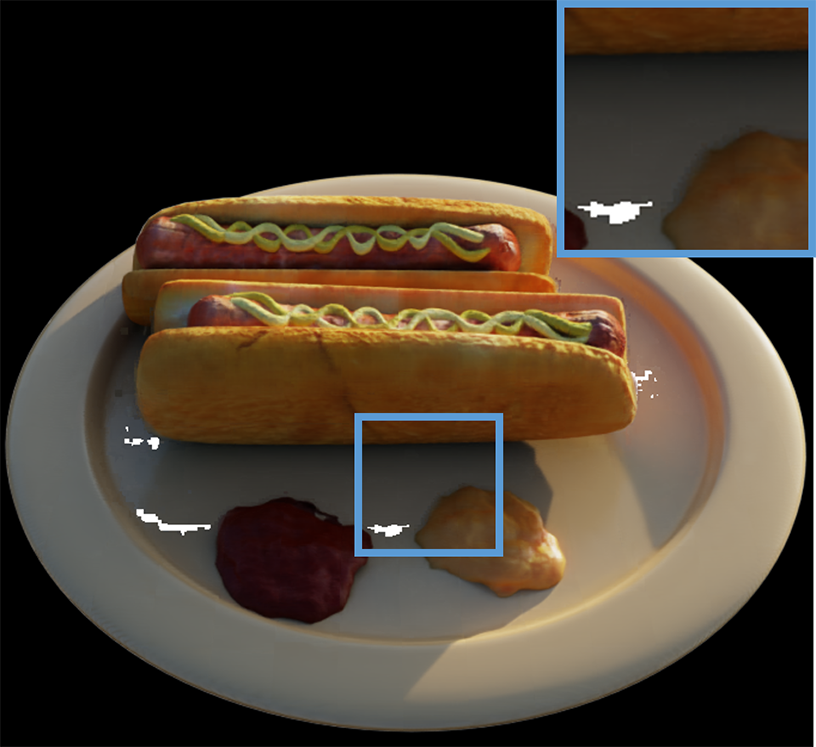} & \includegraphics[align=c,width=0.235\textwidth]{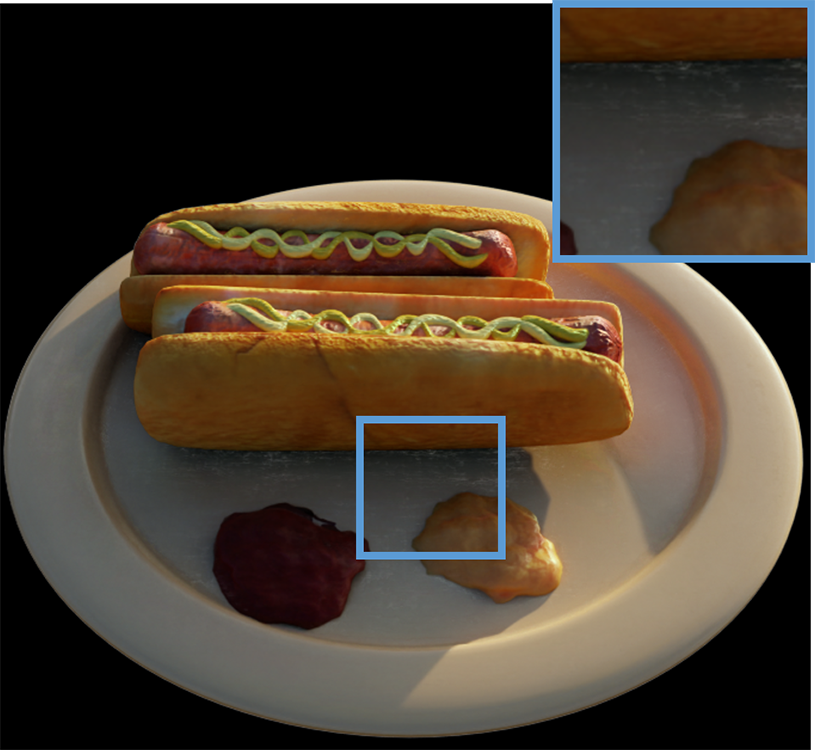} & \includegraphics[align=c,width=0.235\textwidth]{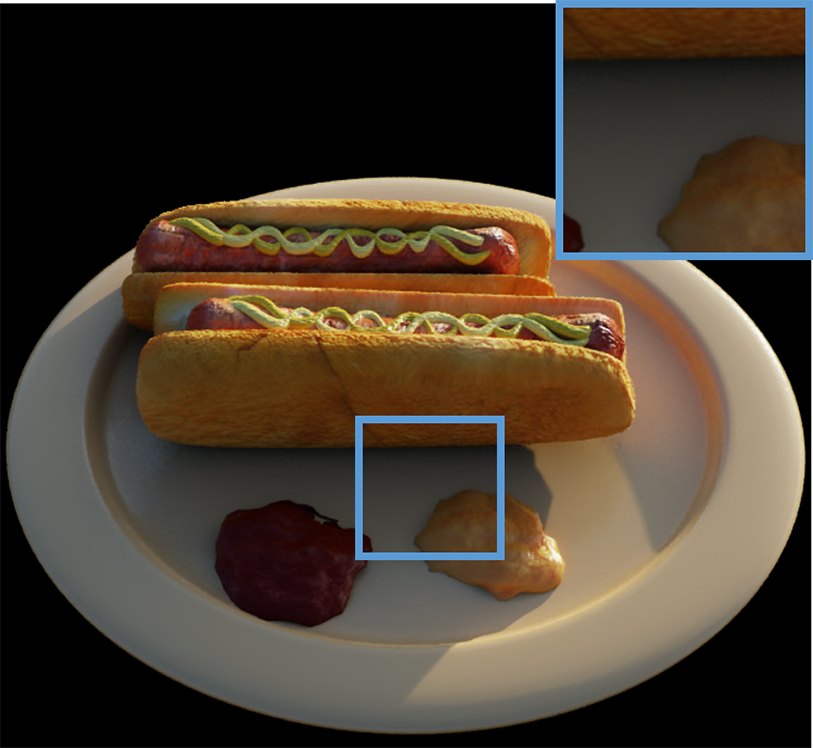} \\

\rotatebox[origin=c]{90}{1920 $\times$ 1080 \strut} & 
\includegraphics[align=c,width=0.235\textwidth]{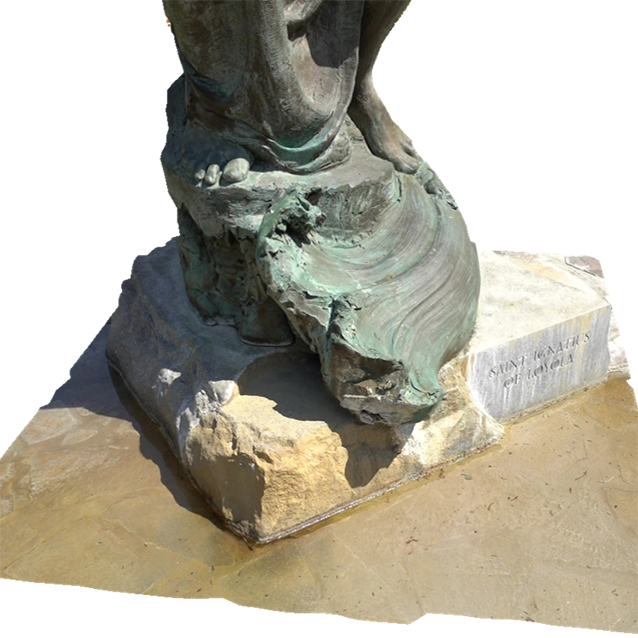} & \includegraphics[align=c,width=0.235\textwidth]{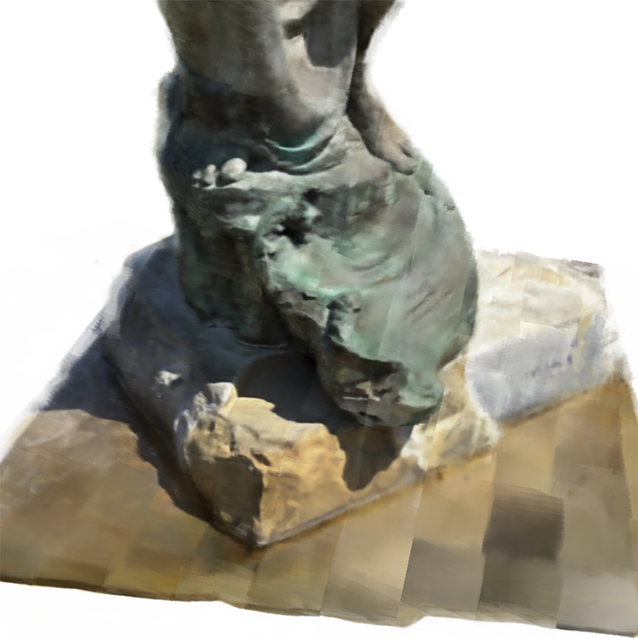} & \includegraphics[align=c,width=0.235\textwidth]{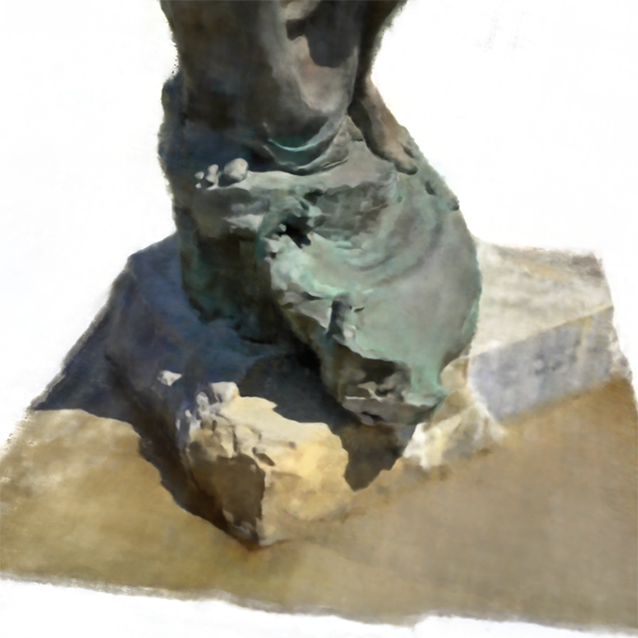} & \includegraphics[align=c,width=0.235\textwidth]{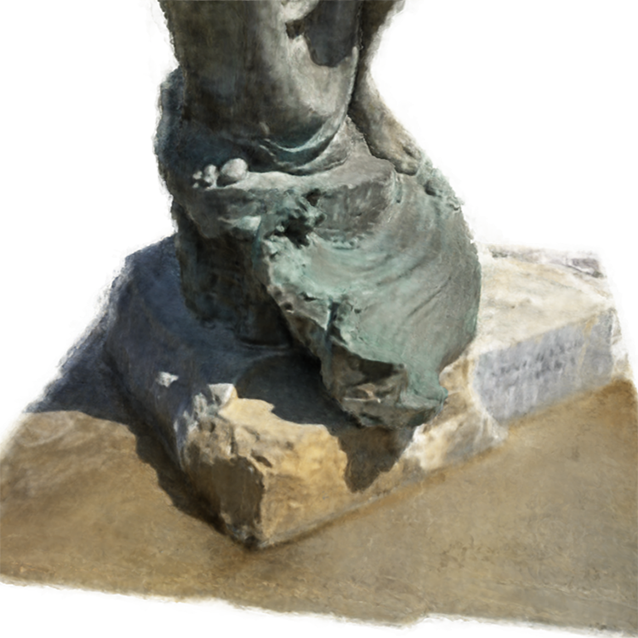} \\

& & & & \\
& Ground Truth & KiloNeRF \cite{reiser2021kilonerf} & SNeRG \cite{hedman2021snerg} & Ours

\end{tabular}
\end{center}
\vspace{-6mm}
\caption{\textbf{Qualitative comparisons with various baselines on Synthetic-NeRF (top) and Tanks\&Temples (bottom) datasets.} We  show the novel views synthesized by KiloNeRF, SNeRG and ours. Our approach effectively speeds up the inference efficiency while maintaining the high quality of the rendered frames. }
\label{fig:comp_sota}
\vspace{-5mm}
\end{figure*}

We provide extensive quantitative comparisons in \cref{tab:quantitative_1}.
Compared with CUDA-based KiloNeRF \cite{reiser2021kilonerf}, our CUDA version achieves significantly better rendering performance in terms of PSNR (more than 1\,dB) and SSIM on the Synthetic-NeRF dataset \cite{mildenhall2020nerf} with much higher run-time FPS and lower training time.
The PSNR of our CUDA-based model on Tanks\&Temples \cite{knapitsch2017tanks} is slightly lower than KiloNeRF, even though the novel views of KiloNeRF contain severe artifacts, as shown in \cref{fig:comp_sota}.
This may be caused by the characteristics of PSNR, which only measures pixel-level differences and ignores the structural similarity.
In this situation, SSIM would be a better metric to indicate the performance gap between our method and KiloNeRF (0.915 vs. 0.910).

\begin{table}
\caption{\label{tab:ablation}%
    \textbf{Quantitative ablation study of proposed components.} Our full model performs significantly better than all variations.}
\vspace{-0.5em}
\centering
\resizebox{\linewidth}{!}{%
\begin{tabular}{ccc|ccc}
\hline
Duplex & Conv & Distill & PSNR ${\uparrow}$                          & SSIM ${\uparrow}$                           & LPIPS ${\downarrow}$                        \\ \hline
      & $\checkmark$    & $\checkmark$       &  26.95                             &       0.884                        &                      0.12        \\
$\checkmark$      &      & $\checkmark$       &                   30.03            &        0.942                       &   0.05                           \\
$\checkmark$      & $\checkmark$    &         &   29.66                            &    0.932                           &                0.06              \\
$\checkmark$      & $\checkmark$    & $\checkmark$       & \cellcolor[HTML]{DAE8FC}32.14 & \cellcolor[HTML]{DAE8FC}0.957 & \cellcolor[HTML]{DAE8FC}0.03 \\
\hline
\end{tabular}%
}
\end{table}

We also compare with WebGL-based SNeRG \cite{hedman2021snerg}.
%
%
Even at 1920$\times$1080 resolution, our WebGL-based model achieves 180+ FPS on a desktop-level GPU, and 30+ FPS on a laptop using Chrome, with better rendering fidelity than SNeRG.
%
In addition, our method requires about two orders of magnitude less computation for training compared to SNeRG. We also show the trade-off comparison regarding memory consumption, image quality and rendering speed in \cref{tab:mem_webgl} under high-resolution setting, which comprehensively demonstrates the superiority of our method.

\paragraph{Qualitative Comparison}

To further verify our method's benefit in terms of
rendering quality over existing fast rendering NeRF techniques, we show qualitative comparisons in \cref{fig:comp_sota}.
As we can see, both baselines, KiloNeRF and SNeRG, cannot render the high-quality details of the scenes, which are possibly bounded by the voxel resolution; however, simply increasing the voxel resolution will directly lead to cubic growth of memory cost.
Besides, some texture artifacts and wrong view-dependent specular highlights also exist in the baselines, which are visually inferior to our method.
More importantly, in the Tanks\&Temples dataset, we could observe vibrant blocking degradation from the results of KiloNeRF.
The potential reason is KiloNeRF has limited generalization capability to out-of-distribution (OOD) views since there are no correlations between space-isolated MLPs.
By contrast, our method can render consistent and high-fidelity views with good view generalization ability.

\subsection{Ablation Study}

\def\swthree{0.25\linewidth}
\setlength{\tabcolsep}{0.5pt}
\begin{figure}[t]
    \begin{center}
    \scriptsize
\begin{tabular}{ccccc}
\includegraphics[width=\swthree]{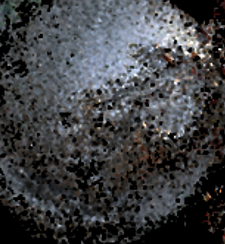}&
\includegraphics[width=\swthree]{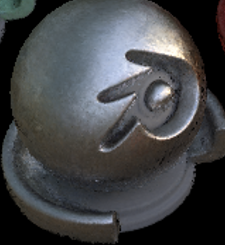}&
\includegraphics[width=\swthree]{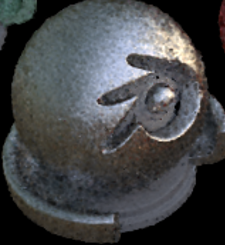}&
\includegraphics[width=\swthree]{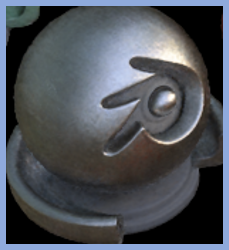}\\
No Duplex&No Conv&No Distill& Full Model
\end{tabular}
\vspace{-3mm}
\caption{\textbf{Qualitative results for each ablation study.} Our full model achieves the best rendering quality among all variations.}
\label{fig: ablation_component}
\end{center}
\vspace{-5mm}
\end{figure}

\paragraph{Duplex Radiance}

To verify the effectiveness of our proposed neural duplex radiance, we compare with a variation that only considers a single ray-surface intersection.
More specifically, we only adopt the mesh extracted from NeRF model with threshold $10^{-4}$ to ensure geometry proxy could cover all pixels and avoid hole artifacts.
As shown in \cref{tab:ablation} and \cref{fig: ablation_component}, rendering without duplex radiance dramatically downgrades the quality of novel-view synthesis since the coarse geometry extracted from NeRF could not provide reasonable shading locations.
More results about using single radiance feature but with different thresholds can be found in the supplementary materials.

\paragraph{Convolution vs MLP}

In contrast to most NeRF methods, which integrate radiance only along a single ray, we aggregate all the radiance information of the local geometry surface via convolution kernels.
\Cref{fig: ablation_component} and \cref{tab:ablation} demonstrate that the convolutional shading network can render high-fidelity novel views more robustly, and effectively overcomes the under-constrained problem while only considering two sample locations for each ray.

\def\swthree{0.25\linewidth}
\setlength{\tabcolsep}{0.5pt}
\begin{figure}[t]
\centering
\scriptsize
\begin{tabular}{ccccc}
\includegraphics[width=\swthree]{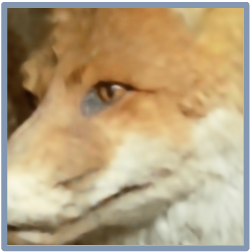}&
\includegraphics[width=\swthree]{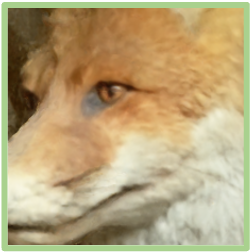}&
\includegraphics[width=\swthree]{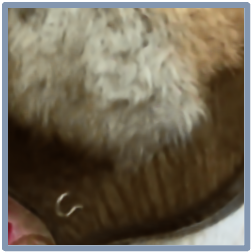}&
\includegraphics[width=\swthree]{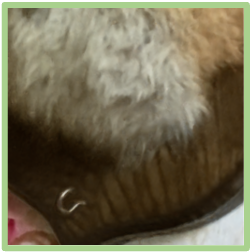}\\
Deeper&Ours&Deeper&Ours
\end{tabular}
\vspace{-3mm}
\caption{\textbf{Ablation study on the network size.} Increasing the depth and receptive field of the convolutional shading network does not lead to an improvement on rendering quality in a consistent manner. Zoom in for more details.}
\label{fig: ablation_on_network_size}
\vspace{-4mm}
\end{figure}

\paragraph{No Distillation}

In this ablation study, we remove all the distillation views and only train the model based on known views.
Our results in \cref{fig: ablation_component} show that optimization without distillation views will lead to high-frequency artifacts in the geometry, which the spatial convolutional kernels are unable to effectively handle as well.

\begin{figure*}
\centering
\small
\def\swthree{0.30\linewidth}
\setlength{\tabcolsep}{5pt}
\begin{tabular}{ccc}
    \includegraphics[width=\swthree]{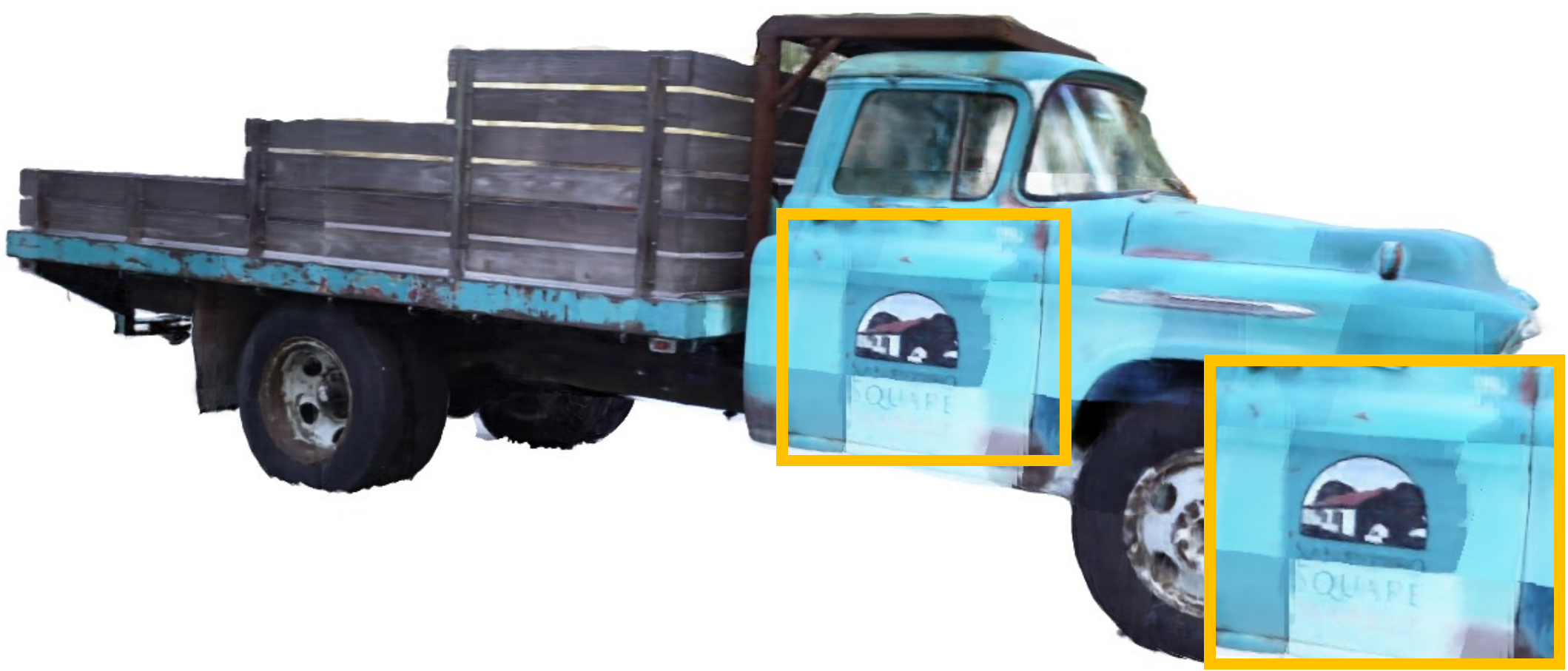} &
    \includegraphics[width=\swthree]{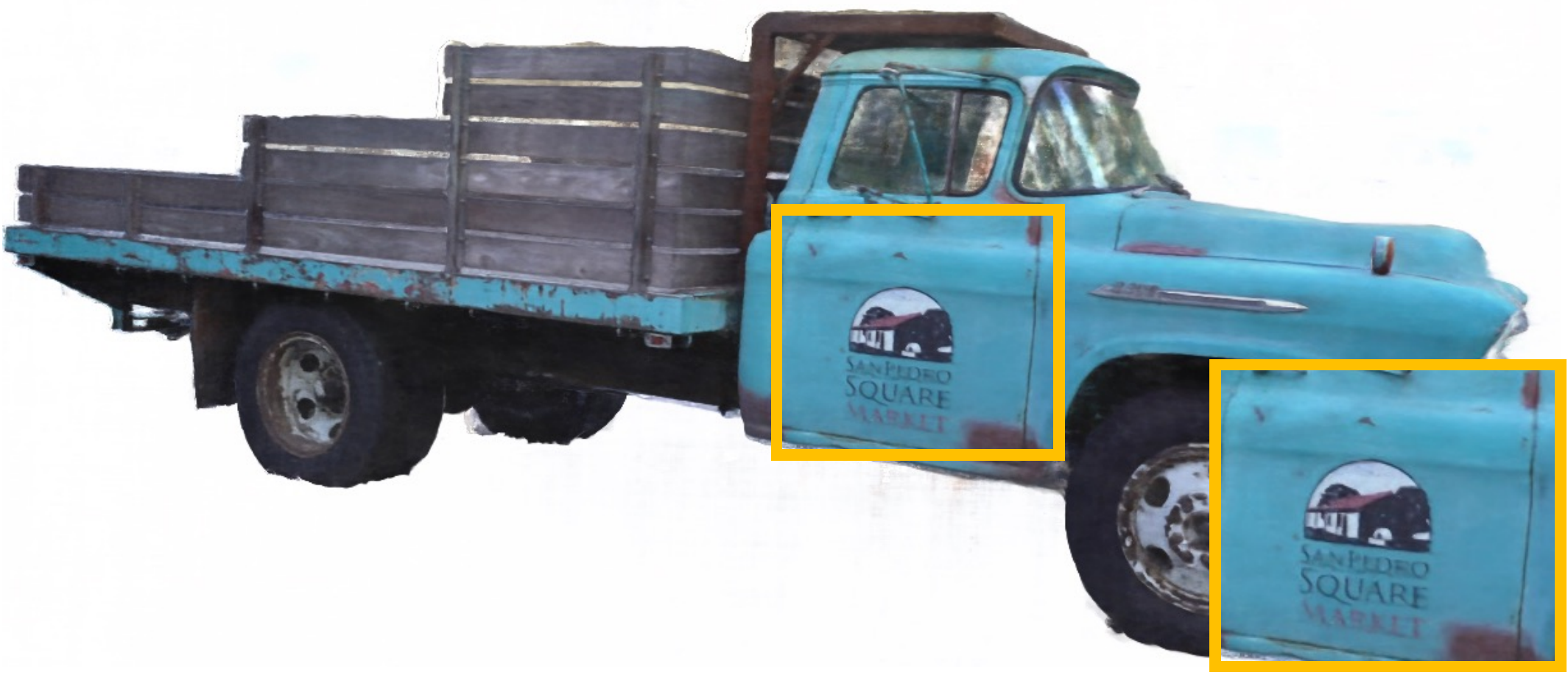} &
    \includegraphics[width=\swthree]{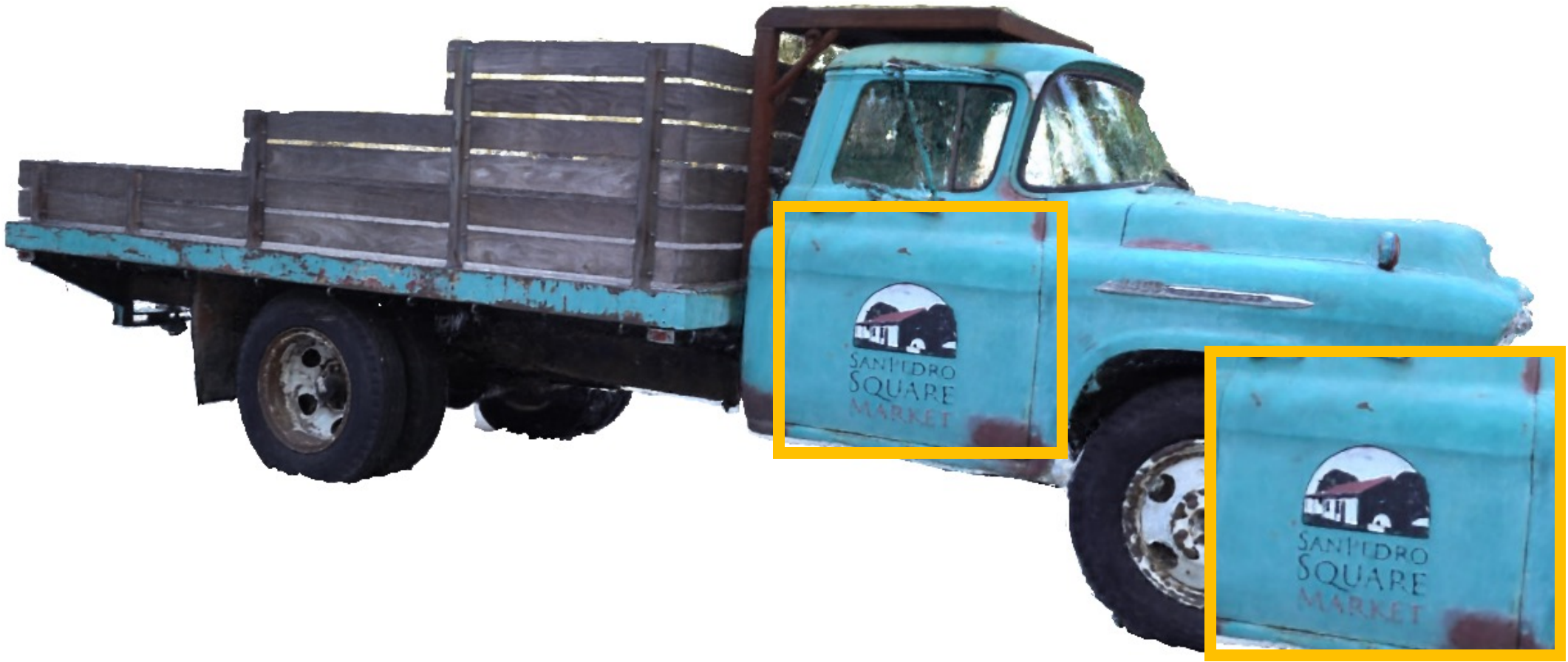} \\
    KiloNeRF~\cite{reiser2021kilonerf} & SNeRG~\cite{hedman2021snerg} & Ours
\end{tabular}
\vspace{-3mm}
\caption{\textbf{Comparison of out-of-distribution (OOD) view synthesis (No ground-truth).} Our method could generate high-quality and view-dependent frames with sharper details for OOD views. }
\label{fig: ablation_OOD}
\vspace{-4mm}
\end{figure*}

\paragraph{Influence of Network Size.}

We have noticed that decreasing the radiance feature dimension and network size will result in some quality degeneration in \cref{tab:quantitative_1}.
Hence, one interesting question is whether performance can be boosted consistently by increasing the network parameters and receptive fields.
We conduct this experiment on the Fox dataset.
More specifically, we double the network layers and boost the original used convolution layers with larger receptive fields.
As shown in \cref{fig: ablation_on_network_size}, the deeper network does not bring more quality gains.
By contrast, the synthesized novel views become more blurry than our method and lose many textural details, which may be caused by the learning difficulty of a deep network.
Leveraging more advanced architectures or loss functions, such as adversarial objectives, could potentially resolve this issue.

\paragraph{Generalization to out-of-distribution views}

NeRF has strong capabilities to interpolate novel views, but its rendering quality may drop for some novel views which don't follow the original distribution.
This issue will be more serious for distillation-based methods that try to mimic the teacher behavior.
We show out-of-distribution (OOD) synthesis results in \cref{fig: ablation_OOD}.
Both KiloNeRF \cite{reiser2021kilonerf} and SNeRG \cite{hedman2021snerg} do not generalize well to OOD views and produce inconsistent and blurry artifacts, while our method synthesizes sharper and cleaner appearance with appropriate lighting.

\begin{table}
\caption{\label{tab:comparison_with_teacher}%
\textbf{Quantitative comparisons with the teacher model} on the Tanks\&Temples~\cite{knapitsch2017tanks} dataset (1920$\times$1080).
}
\vspace{-0.5em}
\centering
\resizebox{\linewidth}{!}{%
\begin{tabular}{lcccr}
\toprule
                                        & PSNR ${\uparrow}\quad$  & SSIM ${\uparrow}\quad$  & LPIPS ${\downarrow}\quad$  & FPS $\uparrow$  \\ \midrule
Original NeRF~\cite{mildenhall2020nerf} &     28.32   &   0.900   &   0.11    &  0.01   \\ 
Teacher TensoRF~\cite{Chen2022ECCV}     &   \bf  28.48   & \bf  0.918   &   0.12    &  0.10   \\ 
Ours w/ WebGL                           &     27.08   &   0.895   &   0.13    &  \bf 186.94   \\
Ours w/ CUDA                            &     28.12   &   0.915   &  \bf 0.09    &   27.49  \\
\hline
\end{tabular}%
}
\end{table}

\paragraph{Comparison with Teacher Model.}

Since our method involves multi-view distillation, we also measure the quantitative differences between our model and the teacher model \cite{Chen2022ECCV} in \cref{tab:comparison_with_teacher}.
Although the numerical results of PSNR and SSIM are slightly inferior to the teacher model, the perceptual metric LPIPS, which better correlates with human perception, surprisingly attains better performance benefiting from the better generalization ability of neural duplex radiance field.
On the other hand, the teacher model TensoRF \cite{Chen2022ECCV} has already optimized the run-time performance by skipping the importance sampling step of original NeRF model and tensor decomposition, but the speed is still limited for real-world interactive applications.
In contrast, our method achieves $\sim$30 FPS while maintaining comparable rendering quality. Significantly, our WebGL version reaches 10,000$\times$ speed up over the original NeRF model.


\paragraph{Comparison with MobileNeRF \cite{ChenFHT2023}}

Recently, a concurrent work MobileNeRF \cite{ChenFHT2023} also explores the rasterization pipeline to accelerate the rendering speed of neural radiance fields.
Nonetheless, their framework involves optimizations of both geometry and appearance 
leading to a high computational cost 
(21+ hours using 8$\times$V100).
By contrast, our method can convert a pretrained NeRF in a few hours using a single GPU.
We provide some qualitative comparisons in \cref{fig: comparison_mobilenerf} as well.
Our approach produces more appealing results and better view-dependent appearance.

\def\swthree{0.25\linewidth}
\setlength{\tabcolsep}{0.5pt}
\begin{figure}[th]
\centering
\scriptsize
\begin{tabular}{ccccc}
\includegraphics[width=\swthree]{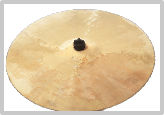}&
\includegraphics[width=\swthree]{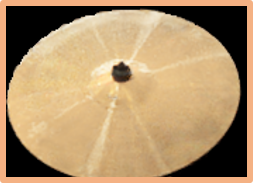}&
\includegraphics[width=\swthree]{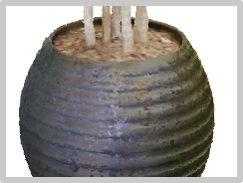}&
\includegraphics[width=\swthree]{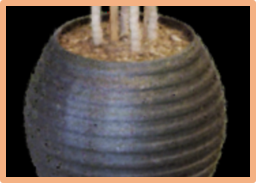}\\
MobileNeRF~\cite{ChenFHT2023}&Ours w/ WebGL&MobileNeRF~\cite{ChenFHT2023}&Ours w/ WebGL
\end{tabular}
\vspace{-3mm}
\caption{\textbf{Qualitative comparisons with MobileNeRF.} Our method produces smoother appearance.}
\label{fig: comparison_mobilenerf}
\end{figure}

%% file: 5-conclusion.tex
\section{Conclusion}

In this paper, we proposed a novel approach for learning a neural duplex radiance field from a pretrained NeRF for real-time novel-view synthesis. 
We employed a convolutional shading network to improve the rendering quality and also proposed a multi-view distillation strategy for better optimization.
%
%
Our method provides significant improvements on run-time, bettering or preserving the rendering quality, and at the same time, operating on a much lower computational cost compared to existing efficient NeRF methods.
%

\paragraph{Limitations} It is challenging to extract useful duplex meshes from NeRFs to represent transparent or semi-trans\-parent scenes and our method is no different.
Integrating order-independent transparency (OIT) with the NeRF framework may be a potential solution to speed up its rendering.
In addition, the quality of the geometry proxy will directly influence the learning of neural representations, thus we would also like to explore scene-specific thresholds rather than empirical defined values while generating meshes, for instance, based on signed distance functions or on the statistics of the density values, for more robust proxy extraction.
Currently, our method focuses on object-level and bounded-scene datasets.
Efficiently rendering real-world unbounded scenes is a promising future direction.
Finally, our experiments demonstrate that sufficient learnable parameters are crucial to achieve better rendering fidelity.
Hence, to counter the performance degradation in the WebGL version, boosting the running efficiency of larger CNN in the fragment shader, or leveraging other acceleration frameworks, could also be considered in future work.

\paragraph{Acknowledgements}
We thank the anonymous reviewers for their constructive comments.
We also appreciate helpful discussions with Feng Liu, Chakravarty R. Alla Chaitanya, Simon Green, Daniel Maskit, Aayush Bansal,  Zhiqin Chen, Vasu Agrawal, Hao Tang, Michael Zollhoefer, Huan Wang, Ayush Saraf and Zhaoyang Lv.
This work was partially supported by a GRF grant (Project No. CityU 11216122) from the Research Grants Council (RGC) of Hong Kong.